\pdfoutput=1

\documentclass[11pt]{article}

\usepackage[preprint]{acl}

\AtBeginDocument{\nolinenumbers}

\usepackage{times}
\usepackage{latexsym}

\usepackage[T1]{fontenc}

\usepackage[utf8]{inputenc}

\usepackage{microtype}

\usepackage{courier}

\usepackage[table]{xcolor}
\usepackage{url}            
\usepackage{booktabs}       
\usepackage{amsfonts}       
\usepackage{graphicx}
\usepackage{listings}
\usepackage{amsmath, amssymb}
\usepackage{multirow}
\usepackage{colortbl}
\usepackage{adjustbox}
\usepackage{pifont}
\usepackage{threeparttable}

\definecolor{lightpurple}{rgb}{0.862, 0.918, 0.992}
\definecolor{aliceblue}{rgb}{0.941, 0.972, 1.000}
\definecolor{peachpuff}{rgb}{1.000, 0.855, 0.725}
\definecolor{mistyrose}{rgb}{1.000, 0.894, 0.882}
\definecolor{seashell}{rgb}{1.000, 0.961, 0.933}

\newcommand{\cmark}{\textcolor{green!50!black}{\ding{51}}}
\newcommand{\xmark}{\textcolor{red!65!black}{\ding{55}}}
\newcommand{\yes}{\makebox[1.05em][c]{\cmark}}
\newcommand{\no}{\makebox[1.05em][c]{\xmark}}

\usepackage[breakable]{tcolorbox}

\newtcolorbox{promptbox}{
  colback=gray!10,
  colframe=gray!50,
  boxrule=0.5pt,
  arc=0pt,
  left=5pt,
  right=5pt,
  top=5pt,
  bottom=5pt,
  breakable
}

\definecolor{mylavender}{RGB}{220,208,255}
\definecolor{mylavenderframe}{RGB}{180,160,220}
\definecolor{pastelyellow}{RGB}{255, 253, 230}

\tcbset{
  mypromptstyle/.style={
    enhanced,
    breakable,
    colback=blue!3!white,
    colframe=blue!80!black,
    coltitle=white,
    colbacktitle=blue!70!black,
    title={\large\bfseries Prompt for Single Choice in OM/ECON/FIN},
    fonttitle=\bfseries\large,
    boxrule=1pt,
    arc=4pt,
    top=8pt,
    bottom=8pt,
    left=8pt,
    right=8pt,
    sharp corners=south,
    attach boxed title to top left={yshift=-2mm, xshift=5mm},
  }
}

\title{\raisebox{-0.35\height}{\includegraphics[height=1.2cm]{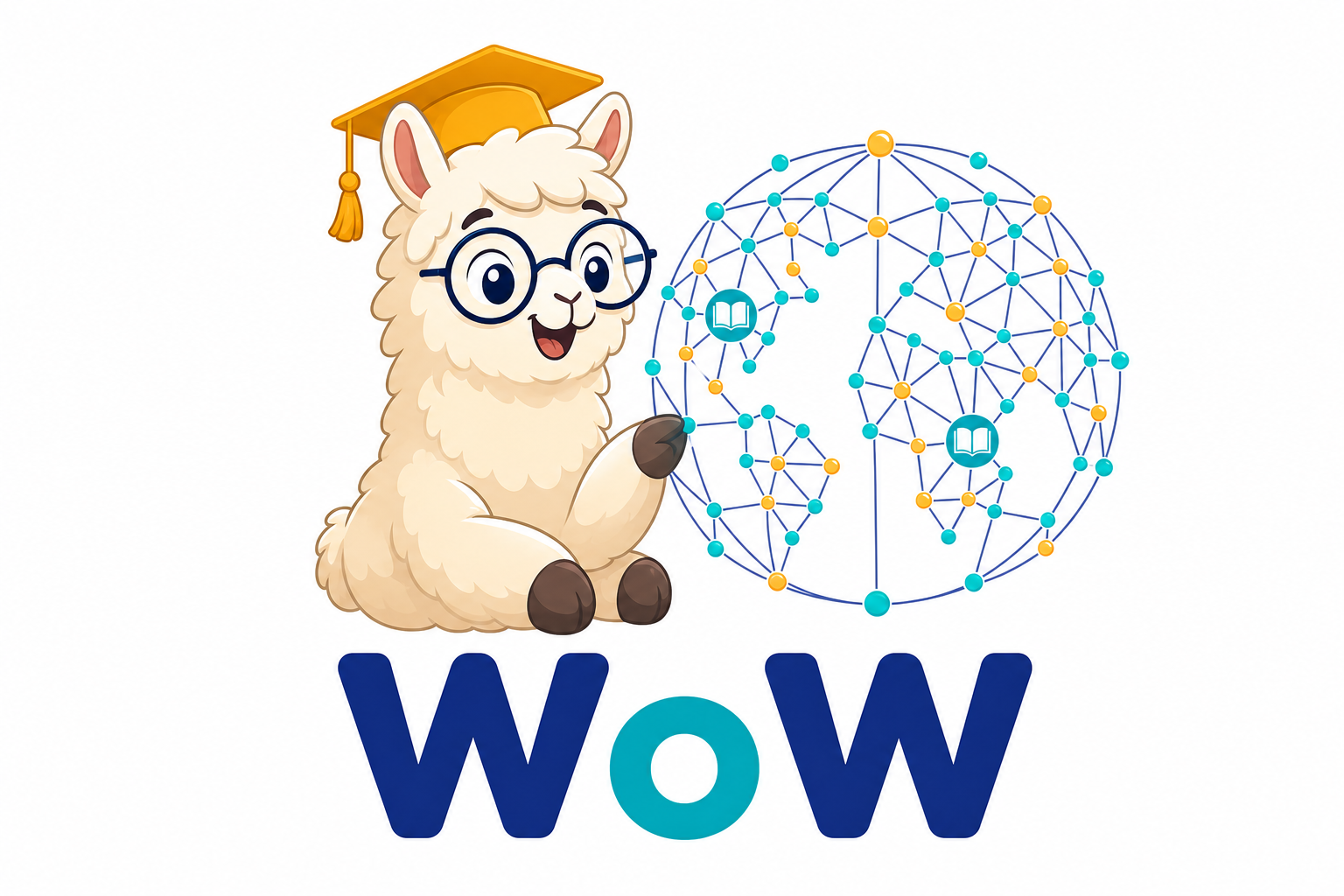}}~The WoW: A Large-Scale World Knowledge Corpus for Full-Lifecycle LLM Training}

\author{
  Tianze Sun \\
  Harbin Institute of Technology
  \And
  Yongyao Duan \\
  OpenCSG
  \AND
  Sam Chen \\
  OpenCSG
  \And
  Yuanjian Xu \\
  Hong Kong University of Science \\
  and Technology (Guangzhou)
}

\begin{document}
\maketitle

\begin{abstract}
Existing Chinese educational corpora often suffer from fragmented coverage and a lack of systematic organization. In this paper, we propose WoW (\textbf{Wo}rld Kno\textbf{w}ledge), a large-scale dataset designed for the full-cycle training of foundation models. The construction of this dataset is driven by an Educational Knowledge Graph, which serves as a taxonomic backbone for the systematic collection and completion of missing educational modules across all levels. Compared to previous versions, this iteration provides enhanced systematicity, superior quality, and a larger scale, while incorporating multimodal signals and supporting every stage of model development.  Experiments demonstrate that WoW significantly improves model performance across diverse benchmarks. To facilitate community research, we further provide an interactive platform for graph-based data selection and retrieval. The link is \url{}.
\end{abstract}
\section{Introduction}


\label{sec:intro}

The knowledge and reasoning ability of large language models (LLMs) is shaped by the data they are trained on at every stage, from pre-training corpora to instruction data and evaluation suites. Small models trained on ``textbook-quality'' data can rival far larger ones \citep{gunasekar2023textbooks, li2023textbooks}, and a few thousand carefully chosen instructions suffice to align a strong base model \citep{zhou2023lima}. Recent corpus efforts have therefore shifted from sheer scale to curation: web corpora are filtered with educational-quality classifiers \citep{penedo2024fineweb, yu2025opencsg}, mixed by explicit data recipes \citep{li2024datacomp, soldaini2024dolma}, mined for mathematical content \citep{paster2024openwebmath, shao2024deepseekmath, azerbayev2024llemma}, or assembled for a single domain such as law \citep{henderson2022pile, niklaus2024multilegalpile}, chemistry \citep{mirza2026chempile}, biomedicine \citep{lozano2025biomedica, touchent2026biomed}, or source code \citep{kocetkov2022stack}. Instruction data has followed a similar path, from aggregated task collections \citep{wang2022super, longpre2023flan} and model-generated instructions \citep{wang2023selfinstruct, xu2024wizardlm, mukherjee2023orca} to instructions synthesized from raw documents \citep{cheng2024adapting, cheng2024instruction, yue2024mammoth2} and domain-specific collections for scientific reasoning \citep{zhang2024sciinstruct}, molecular tasks \citep{fang2024mol, yu2024llasmol}, and textbook-grounded tutoring \citep{wang2024book2dial, dan2023educhat}. A growing family of benchmarks measures disciplinary knowledge in English and Chinese \citep{hendrycks2020measuring, huang2023ceval, li2023cmmlu, wang2023scibench, sun2024scieval}.

Educational data, and Chinese educational data in particular, nevertheless remains difficult to use systematically. We identify three gaps. \emph{First}, coverage is organized by source rather than by knowledge. Corpora are collected per website, per publisher, or per discipline, and are indexed by quality scores or source labels (Table~\ref{tab:dataset_comparison_main}); without a model of what a curriculum should contain, missing modules, cross-disciplinary links, and prerequisite structure are invisible, so gaps can be neither measured nor filled \citep{zhao2022edukg, liang2026k12}. \emph{Second}, document-derived supervision trades off grounding against difficulty. Question--answer pairs extracted from documents tend to restate isolated facts or turn passages into reading-comprehension exercises \citep{cheng2024adapting}, whereas freely generated problems drift away from their sources and become under-specified; neither yields instances that are at once faithful to the source, self-contained, and demanding enough to elicit reasoning. Imitating the surface form of strong models' answers without the underlying knowledge, moreover, improves style more than capability \citep{gudibande2023false}. \emph{Third}, resources are fragmented across the training lifecycle: pre-training corpora, instruction data, and benchmarks are built independently, with different taxonomies and little provenance linking them, which makes it hard to trace a model's behavior back to the knowledge it was trained on.

We address these gaps with \textbf{WoW} (\textbf{Wo}rld kno\textbf{W}ledge), a large-scale corpus for full-lifecycle LLM training built around an Educational Knowledge Graph. The graph serves as a taxonomic backbone: it organizes domains, subjects, and fine-grained KnowledgePoints under a unified ontology aligned with national discipline standards \citep{gbt13745_2009}, maps resources from different stages and sources into a shared semantic space, and exposes under-covered modules that guide further collection (Appendix~\ref{app:kg}). On top of this backbone, a four-phase construction pipeline (\S\ref{sec:method}) turns heterogeneous documents into supervision. Phase~I recovers coherent passages and abstracts them into knowledge units---key equations, central propositions, and result tables, with tightly coupled objects kept together as macro units---and retrieves a sufficient evidence set for each. Phase~II composes each grounded unit into a multi-step research case whose subproblems depend on one another. Phase~III filters candidates with hard synthesis, transformation, and dependency tests and scores them for reasoning depth, knowledge synthesis, and centrality. Phase~IV diversifies the response format while preserving the reasoning signal. Every instance keeps document-level provenance.

Applied to fourteen source collections spanning computer science, the natural sciences, engineering, economics and management, law, and the humanities and social sciences, the pipeline processes 309K documents, abstracts 2.1M knowledge units, and releases 188K instruction instances totaling 146M tokens (\S\ref{sec:data}).

We fine-tune three base models of different strength---Llama-3.2-3B, Llama-3.1-8B \citep{grattafiori2024llama3}, and Qwen3-4B-Base \citep{yang2025qwen3}---on this data with a single, conservative recipe, and evaluate them on a held-out split of documents unseen in training and on eight external benchmarks. Before training, we correct two format artifacts of synthetic selected-response supervision. Rationales that open with the answer teach models to commit before reasoning, forfeiting the benefit of chain-of-thought \citep{wei2022chain}, and skewed answer counts and positions in multiple-choice items induce systematic biases. We therefore rewrite the same records to reason before concluding, rebalance their answer distributions, and shuffle their options. Fine-tuning improves held-out accuracy for every model, by 15.9, 5.2, and 2.8 points respectively, almost entirely through multiple-answer problems that require judging each option against a formal specification. The gain shrinks as the base model grows stronger, and the fine-tuned 3B model surpasses the 8B base model. On external benchmarks, the benefit concentrates where the data is concentrated. The 3B model improves on SciEval by 5.5 points, driven by its chemistry ($+7.3$) and biology ($+6.4$) portions, and on ScienceQA and MoleculeQA, while the larger models gain slightly on science benchmarks. Knowledge outside the scope of WoW is largely preserved: the stronger models stay within 0.7 points of their base checkpoints on broad-coverage exams (MMLU, CMMLU, C-Eval) and on an out-of-domain medical control (MedMCQA). Consistent with prior observations of forgetting under continual fine-tuning \citep{luo2023empirical}, only the smallest model pays a cost, of 1.1--2.6 points.

Our contributions are as follows:
\begin{itemize}
    \item We present WoW, an educational corpus whose collection and completion are driven by an Educational Knowledge Graph, together with an interactive platform for graph-based data selection and retrieval.
    \item We propose a four-phase, provenance-preserving pipeline that converts heterogeneous documents into grounded, self-contained, and reasoning-intensive supervision, and we release 188K instances built from 309K documents.
    \item We provide a detailed analysis of the corpus, including pipeline yield, the composition of knowledge units and cognitive skills, and quality-score distributions.
    \item We fine-tune three base models on WoW and show consistent gains on held-out disciplinary problems, gains on aligned science benchmarks, and preserved broad knowledge for 4B--8B models, using an evaluation protocol that does not depend on the chat format. 
\end{itemize}

\section{Data Construction Pipeline}
\label{sec:method}

\subsection{Overview}
\label{sec:method:overview}

Our goal is to transform heterogeneous educational documents into challenging and verifiable supervision. A useful training instance must satisfy three properties. It must be grounded in the source, self-contained for a reader who has not seen the source, and sufficiently demanding to elicit reasoning rather than retrieval. These requirements are coupled. Increasing difficulty without improving context creates underspecified questions, while adding context without controlling question design produces trivial extraction tasks.

We address this tension with a four-phase pipeline. Given a source document $d$, context generation produces a structured knowledge unit $k$ and an evidence set $E$ that supports it. Problem generation maps $(k,E)$ to a question and reference answer $(q,a)$. Quality control retains only candidates that satisfy grounding and reasoning constraints. The final phase chooses an appropriate response format and converts the retained instances into training records. The overall transformation is
\begin{equation}
\begin{aligned}
 d &\xrightarrow{\text{context}} (k,E)
 \xrightarrow{\text{generation}} (q,a),\\
 (q,a) &\xrightarrow{\text{quality}} (q^{*},a^{*})
 \xrightarrow{\text{formatting}} x .
\end{aligned}
\end{equation}
All operations preserve document-level provenance, which allows every retained statement to be traced to its source. Domain adaptation is introduced through domain-specific concept definitions, cognitive skills, and demonstrations, while the four-phase procedure remains unchanged.

\subsection{Phase I: Context Generation}
\label{sec:method:context}

The first phase converts an unstructured document into the minimal context needed to formulate and solve a rigorous problem. It has two objectives. The first is to recover coherent semantic units from noisy and heterogeneous sources. The second is to identify assessable knowledge and retrieve sufficient evidence for it.

\paragraph{Coherence-Aware Segmentation.}
\label{sec:method:s1}
\label{sec:method:s1_5}

We first normalize PDF, HTML, Markdown, and article archives into a common document representation. \textbf{[What is the common document representation? Define its schema.]} Layout-aware parsing \textbf{[What is layout-aware parsing? How is layout used?]} preserves headings, equations, tables, and their reading order, while deterministic cleaning removes encoding artifacts, boilerplate, and duplicate passages. Low-information articles are filtered using source metadata and content length before model inference. \textbf{[What metadata? Length threshold?]}

Document headings provide useful but imperfect segmentation. A heading boundary may separate an assumption from the proposition it qualifies, whereas a long section may contain several independent arguments. We therefore use headings only to obtain an initial sequence of chunks
\[
\mathcal{C}=(c_1,c_2,\ldots,c_N).
\]
Each chunk is summarized into a compact semantic descriptor $s_i$. For every adjacent pair, a language model predicts a coherence relation
\[
r_i = J(s_i,s_{i+1}), \qquad r_i\in\{0,1\},
\]
where $r_i=1$ indicates that the two chunks should be interpreted as one unit. Maximal contiguous runs connected by positive relations are merged. \textbf{[Example? e.g.\ merge a definition with the claim it scopes ($r{=}1$); keep a liability section separate ($r{=}0$).]} This operation restores local dependencies while preventing unrelated sections from being combined. Structurally obligatory but pedagogically uninformative sections, such as references and administrative statements, are removed before consolidation.

The resulting representation is a sequence of self-contained passages rather than a sequence determined solely by document layout. This distinction is important because all subsequent extraction and generation operate independently on these passages.

\paragraph{Knowledge Abstraction.}
\label{sec:method:s2}

From the consolidated passages, we extract knowledge units that can support explicit assessment. Each unit is centered on one of three source objects: a key equation, a central proposition, or a result table. These objects cover formal mechanisms, testable claims, and empirical evidence while excluding intermediate algebra, peripheral claims, descriptive tables, and routine robustness checks.

An isolated object is not always an adequate unit of reasoning. For example, a proposition may depend on an assumption set and a defining equation, and a regression table may depend on an identification specification. We therefore preserve tightly coupled objects as one macro unit. Formally, a knowledge unit is
\[
k_i=(O_i,h_i,z_i,T_i),
\]
where $O_i$ is an atomic object or an ordered set of dependent objects, $h_i$ is its source section, $z_i$ is a concise semantic description, and $T_i$ is a set of cognitive skills supported by the source. Typical skills include derivation, interpretation, causal analysis, and assumption critique. Duplicate objects are retained only at their first defining occurrence.

This representation separates content from assessment intent. The source objects determine what may be asked, while the skill set determines the reasoning expected from the learner. Domain-specific ontologies instantiate both components without changing the extraction procedure.

\paragraph{Query Intent and Evidence Grounding.}
\label{sec:method:s3}
\label{sec:method:s4}

For each $k_i$, we first construct a query intent that integrates all skills in $T_i$. The skills are not expanded into independent questions. Instead, they are arranged as a dependency chain. A derivation can establish a result, an interpretation can explain its meaning, and an assumption critique can determine when the result remains valid. This produces a single coherent assessment objective rather than a collection of parallel prompts.

We then retrieve evidence directly from the source document. The evidence set contains the passages that establish the focal result and the supporting material required to understand it. Supporting material includes definitions of named concepts, meanings of symbols, assumptions, institutional settings, and any table or mapping used by the intended solution. References to other sections or appendices are followed recursively.

Evidence retrieval is constrained by sufficiency rather than lexical relevance alone. Let $\Gamma(q)$ denote the set of concepts, symbols, assumptions, and operations required by a query intent. A valid evidence set must satisfy
\begin{equation}
\forall g\in\Gamma(q),\quad \operatorname{support}(g,E)=1.
\label{eq:evidence_coverage}
\end{equation}
The retriever is restricted to source selection and may not introduce paraphrased facts. If Equation~\ref{eq:evidence_coverage} is not satisfied, retrieval continues across the document. If the source does not define a required element, the query intent is rejected. Phase I therefore outputs not only relevant context, but a certificate that the planned problem can be made self-contained.

\subsection{Phase II: Problem Generation}
\label{sec:method:s5}

Problem generation turns each grounded pair $(k,E)$ into a compact research case. The generated problem contains a contextual setup, a formal specification, and a sequence of questions. The setup introduces the scientific or technical setting. The specification presents every equation, assumption, table, and symbol needed for a solution. The questions then transform this material into an escalating reasoning process.

We impose a dependency structure on the questions. If $q_j$ denotes the $j$th subproblem, then for every $j>1$, solving $q_j$ must require either the result of an earlier subproblem or a combination of evidence not previously used. This rule excludes lists of independent questions that merely share a topic. At least one subproblem must require an explicit derivation or an equivalently rigorous analytical argument. The final subproblem applies the established reasoning to a counterfactual condition, a parameter change, an identification challenge, or a synthesis of theory and evidence.

The reference answer is generated from the same evidence and follows the dependency order of the questions. Claims copied from the source remain directly grounded in $E$. Claims required by a new scenario must be derivable from the supplied specification. Thus, novelty is introduced through inference over source-supported premises, not through the introduction of unsupported knowledge.

Before a candidate leaves this phase, a completeness audit checks that every symbol and requested concept is defined before use, every numerical operation has the necessary values, and every lookup operation has its mapping. Candidates with unresolved dependencies are revised or discarded.

\subsection{Phase III: Quality Control}
\label{sec:method:s6}

Generation is intentionally permissive, while selection is conservative. We first group candidates by the major argumentative threads of their source document. Compatible candidates are consolidated when they share assumptions and form a stronger reasoning chain. Candidates are kept separate when consolidation would mix incompatible models or create several unrelated difficulty peaks.

Each candidate must pass three hard tests. The synthesis test requires the solution to combine at least two distinct pieces of information. The transformation test requires reasoning beyond quotation or paraphrase. The dependency test requires later subproblems to build on earlier results. Failure on any test removes the candidate.

Candidates that pass the hard tests are scored for reasoning depth $A$, knowledge synthesis $B$, and conceptual centrality $C$, each on a ten-point scale. We compute
\begin{equation}
Q=0.4A+0.4B+0.2C.
\end{equation}
We retain candidates only when $A\geq6$, $B\geq6$, and $Q\geq7$. Near-duplicate candidates are resolved in favor of the higher-scoring and more central problem. This stage produces two broad families: problems that require quantitative reasoning over tables and problems that emphasize derivation, interpretation, or conceptual analysis.

\subsection{Phase IV: Diversification and Formatting}
\label{sec:method:formatting}
\label{sec:method:s7}
\label{sec:method:s8}

The final phase adapts response format without weakening the reasoning signal. Open-ended form is the default because it preserves intermediate reasoning. A problem is converted to selected response only when its conclusion is unambiguous and its common failure modes admit diagnostic distractors. We score these two properties as answer uniqueness $U$ and distractor quality $D$, and permit conversion only when
\[
\tfrac{1}{2}(U+D)\geq9.
\]

Selected-response distractors are derived from plausible conceptual reversals, omitted reasoning steps, sign or unit errors, and misuse of related formulas. They are not generated as arbitrary false statements. The full context and formal specification are retained so that conversion changes only the response interface, not the knowledge available to the learner.

Finally, each retained instance is serialized in both conversational and instruction-oriented schemas. Selected-response answers may include a concise explanation of the correct option and the error represented by each distractor. Samples are shuffled with a fixed seed and divided into training and validation subsets at the document level. The released records retain source provenance, question type, and quality score, supporting both reproducible filtering and downstream analysis.

\section{Data Statistics and Analysis}
\label{sec:data}

\begin{table*}[t]
\centering
\caption{Detailed composition of WoW. \textbf{Discipline} and \textbf{Subject} report the dominant Domain and the two most frequent Subjects assigned by our Educational Knowledge Graph (EKG) to each collection. \textbf{Size} gives the raw source volume and the number of retained supervised instances. Collections marked $^{\star}$ were acquired along EKG Domain and Subject partitions to complete under-covered modules.}
\label{tab:data-stats}
\small
\resizebox{\textwidth}{!}{%
\begin{tabular}{llllll}
\toprule
\textbf{Discipline} & \textbf{Subject} & \textbf{Type} & \textbf{Format} & \textbf{Size} & \textbf{Source (KG-driven)} \\
\midrule
Computer Sci. & Software; Applications & Web article & JSONL & 45GB / 119.8K & CSDN technical community \\
Computer Sci. & Software; Applications & Curated notes & Markdown & 1.6GB / 6.5K & Disciplinary Knowledge Repository (CS)$^{\star}$ \\
Computer Sci. & Software; Architecture & Book/Textbook & PDF$\rightarrow$MD+Img & 1.5TB / 8.1K & High-Quality Structured Documents (CS) \\
Computer Sci. & Software; Applications & Book & PDF & 19GB / 4.1K & Publishing House of Electronics Industry \\
Computer Sci. & Software; Applications & Book & PDF & 3.5GB / 0.2K & Posts \& Telecom Press \\
Engineering & Mechanical; Electrical \& Electronics & Book & PDF & 196GB / 1.7K & China Machine Press \\
Engineering / Econ. & Civil \& Architectural; Applied Econ. & Textbook & PDF & 124GB / 3.8K & Peking University Press e-Textbooks \\
Natural Sci. & Mathematics; Biology & Book/Textbook & PDF$\rightarrow$MD+Img & 592GB / 7.4K & High-Quality Structured Documents (NatSci) \\
Natural Sci. & Astronomy \& Earth Sci.; Physics & Monograph & PDF & 6.1GB / 0.2K & Science Press \\
Multi-disc. & Mathematics; Computer Software & Book & PDF & 41GB / 0.3K & Tsinghua University Press \\
Econ. \& Mgmt. & Applied Econ.; Business Admin. & Book & Markdown & 3.8GB / 2.1K & Disciplinary Knowledge Repository (Econ.)$^{\star}$ \\
Humanities \& Soc. & Philosophy \& History; Sociology \& Pol. Sci. & Book/Textbook & PDF$\rightarrow$MD+Img & 161GB / 4.7K & High-Quality Structured Documents (Soc.) \\
Humanities / Soc. Edu. & Philosophy \& History; Cultural Aesthetics & K--12 Textbook/Book & Markdown & 4.3GB / 10.8K & Disciplinary Knowledge Repository (Social)$^{\star}$ \\
Law & Legal Systems \& Practice; Specific Legal Branches & Statute/Regulation & Markdown & 0.7GB / 19.0K & Disciplinary Knowledge Repository (Law)$^{\star}$ \\
\midrule
\textbf{Total} & 53 Subjects & -- & -- & 2.7TB / 188.7K & 14 collections, 100.5K source documents \\
\bottomrule
\end{tabular}}
\end{table*}

\begin{table}[t]
\centering
\caption{Domain-level coverage of the EKG by WoW. A KnowledgePoint (KP) is covered if at least one instance is attached to it.}
\label{tab:kg-coverage}
\small
\resizebox{\columnwidth}{!}{%
\begin{tabular}{lrrr}
\toprule
\textbf{Domain} & \textbf{KPs} & \textbf{Covered KPs} & \textbf{Instances} \\
\midrule
Computer Science \& Technology & 3,941 & 3,492 (88.6\%) & 135,139 (71.6\%) \\
Law \& Institutional Studies & 10,137 & 2,882 (28.4\%) & 14,088 (7.5\%) \\
Natural Sciences \& Technology & 14,305 & 4,126 (28.8\%) & 10,036 (5.3\%) \\
Humanities \& Social Sciences & 11,523 & 3,636 (31.6\%) & 9,858 (5.2\%) \\
Engineering \& Technology & 7,162 & 2,320 (32.4\%) & 8,230 (4.4\%) \\
Social Edu. \& Literacy & 14,409 & 3,031 (21.0\%) & 6,298 (3.3\%) \\
Economics \& Management & 3,776 & 1,373 (36.4\%) & 3,627 (1.9\%) \\
Medical \& Health Sciences & 11,068 & 861 (7.8\%) & 1,404 (0.7\%) \\
Unclassified \& General KB & 43 & 10 (23.3\%) & 13 (0.0\%) \\
\midrule
\textbf{Total} & 76,364 & 21,731 (28.5\%) & 188,693 \\
\bottomrule
\end{tabular}}
\end{table}

\paragraph{Scale and Composition.}
Table~\ref{tab:data-stats} summarizes the composition of WoW. The corpus is built from about 2.7\,TB of raw educational material covering web articles, textbooks, monographs, disciplinary repositories, and legal texts. Applying the pipeline to these sources yields 188.7K supervised instances grounded in 100.5K source documents. Selection is deliberately conservative. Most documents contribute only the single problem that passes every grounding and reasoning test, which favors depth of supervision over volume. Each instance keeps its document-level provenance and its assigned KnowledgePoint, so it can be traced both to the text that supports it and to its position in the graph.

\paragraph{Knowledge-Graph Organization.}
WoW is organized by the EKG, not by source. The graph spans 9 Domains, 55 Subjects, and 76,364 KnowledgePoints. They are connected by 318K typed edges, of which lateral (\texttt{RELATED\_TO}), prerequisite (\texttt{PRECURSOR\_TO}), and contrastive (\texttt{CONTRAST\_TO}) relations make up more than 70\%. Instances cover 53 Subjects and 21,731 KnowledgePoints (Table~\ref{tab:kg-coverage}). This organization matters because no single source corresponds to a single discipline. Even an individual publisher's catalog spreads across dozens of Subjects. A textbook series may therefore contribute to civil engineering, applied economics, and business administration at the same time. Grouping instances by KnowledgePoint brings together material on the same concept from different sources and media. A source-based layout would scatter that material.

\paragraph{Instance Characteristics.}
Open-ended analytical problems make up 69.6\% of instances. The rest are single-choice (15.6\%), multiple-choice (12.7\%), and table-grounded quantitative problems (2.1\%). Selected-response formats appear mostly in shorter, self-contained technical material. Problems derived from textbooks and monographs are almost always open-ended, because their conclusions seldom satisfy the uniqueness and distractor-quality criteria needed for conversion. Because every instance must be self-contained, instances are long. The average prompt, which includes the setup, the formal specification, and the dependent subquestions, has 2.0K characters, and the average reference answer has 1.5K. Book-derived problems are about twice as long as those from web articles (3.5K versus 1.6K characters per prompt), reflecting the larger evidence sets needed to ground reasoning in textbook chapters. Source passages stay in their original language and all metadata is normalized to English, which makes WoW a bilingual Chinese--English resource.

\paragraph{Coverage Analysis.}
The graph turns corpus balance into a quantity we can measure. Computer science is the most densely covered Domain: 88.6\% of its KnowledgePoints have attached instances. Coverage of other Domains is much sparser, and Medical \& Health Sciences is the sparsest at 7.8\%. This per-node view supports targeted rather than opportunistic data collection. Under-covered Domains and Subjects can be identified directly, and new sources can be sought along the corresponding graph partitions. Several collections in WoW were obtained this way, including the economics, social-science, and legal repositories, and they increased coverage of their Domains. Because coverage is defined over KnowledgePoints rather than over sources, the contribution of any new collection can be stated precisely as the set of graph nodes it fills.
\section{Experiments}
\label{sec:experiments}

We test whether supervision built by the pipeline improves models in the domains it covers, and what it costs elsewhere. We fine-tune three base models of different strength on the instruction subset of WoW and evaluate them on the held-out split of WoW and on eight external benchmarks drawn from Table~\ref{tab:dataset_comparison_main}.

\subsection{Setup}
\label{sec:exp-setup}

\paragraph{Models and training.}
We fine-tune Llama-3.2-3B, Llama-3.1-8B \citep{grattafiori2024llama3}, and Qwen3-4B-Base \citep{yang2025qwen3} with full-parameter supervised fine-tuning on 183K training instances (148.6M tokens with the Llama tokenizer). All three use the same recipe: one epoch, learning rate $5\times10^{-6}$ with cosine decay, an effective batch of 128 sequences of up to 4,096 tokens, and a loss computed only on response tokens. Training data, prompts, and evaluation items share the chat template of each model, with thinking disabled for Qwen3 and no system prompt, so that models are trained under the same conditions in which they are tested. Appendix~\ref{app:settings} gives the full configuration.

\paragraph{Training data.}
Models are trained on the released instances after the cleaning described in \S\ref{sec:data} and two format corrections of the selected-response items, both of which rewrite existing records without regenerating them. First, each answer states its reasoning before its conclusion: in the raw records, the chosen option opens the answer and the justification follows, which teaches models to commit before reasoning. Second, answer distributions are rebalanced. One quarter of the multiple-answer items now have a single correct option, obtained by re-framing single-answer items as ``select all that apply,'' since their distractors are false by construction. Options are also shuffled so that the correct letter is uniformly distributed (it was B in 64\% of the raw single-answer items). Training and validation splits are drawn at the document level, so no document contributes to both training and evaluation.

\paragraph{Evaluation.}
\emph{Held-out split.} The 3,175 test instances come from 1,688 documents unseen in training. We report accuracy on its 881 selected-response items (486 single-answer, 395 multiple-answer) under exact match of the predicted option set, and F1 for the multiple-answer items. Base and fine-tuned models are evaluated identically: each question is given as plain text after three exemplars of the same question type drawn from the training split, which base models can follow without a chat format.
\emph{External benchmarks.} We evaluate knowledge with the base-model protocol of lm-evaluation-harness \citep{gao2024lmeval}: plain-text prompts, the benchmark's own few-shot format, and prediction by the highest next-token probability among the option letters. This protocol does not depend on the chat template, so base and fine-tuned models are compared on equal terms. We use broad-coverage exams in English and Chinese (MMLU, CMMLU, C-Eval; 5-shot), science benchmarks aligned with our sources (ChemBench, ScienceQA, SciEval, MoleculeQA; 0-shot), and MedMCQA (0-shot) as an out-of-domain control, since our sources contain almost no medicine. For every comparison we report the paired difference between base and fine-tuned model with a 95\% bootstrap confidence interval over items.

\subsection{Results on the Held-out Split}

\begin{figure*}[t]
\centering
\includegraphics[width=\textwidth]{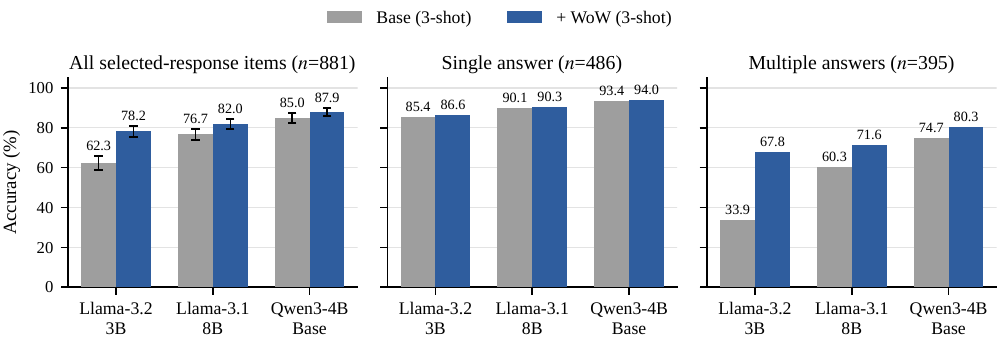}
\caption{Selected-response accuracy on the held-out split of WoW. Error bars in the left panel are 95\% bootstrap confidence intervals. Fine-tuning gains over the base model (paired, 95\% CI): $+15.9$ [12.6, 19.3] for Llama-3.2-3B, $+5.2$ [2.4, 8.2] for Llama-3.1-8B, and $+2.8$ [0.8, 5.0] for Qwen3-4B-Base; all gains come from multiple-answer items.}
\label{fig:heldout}
\end{figure*}

Figure~\ref{fig:heldout} shows that fine-tuning improves held-out accuracy for all three models, with paired gains of $+15.9$ points for Llama-3.2-3B, $+5.2$ for Llama-3.1-8B, and $+2.8$ for Qwen3-4B-Base; all three confidence intervals exclude zero. Two regularities stand out. First, the gain shrinks as the base model becomes stronger. The fine-tuned 3B model (78.2\%) surpasses the 8B base model (76.7\%), whereas Qwen3-4B-Base already solves 85.0\% of the items before fine-tuning. Second, the gains come almost entirely from multiple-answer items ($+33.9$, $+11.4$, and $+5.6$ points). Single-answer accuracy changes by at most 1.2 points, and no single-answer change is significant. Multiple-answer items require judging every option against the problem's specification rather than recognizing the most plausible one, and this is the skill that the reasoning-first answers in WoW demonstrate. Multiple-answer F1 rises accordingly, from 82.5 to 91.1, 89.7 to 92.4, and 92.8 to 95.4.

\subsection{Results on External Benchmarks}

\begin{table*}[t]
\centering
\caption{Knowledge evaluation on external benchmarks (accuracy, \%; plain-text prompts, next-token prediction over option letters). MMLU, CMMLU, and C-Eval are 5-shot; the others are 0-shot. ChemBench is restricted to its 2,147 single-answer items, ScienceQA to its 2,224 text-only test items, and SciEval to the 10,704 multiple-choice items of its test set. $\Delta$: difference to the base model; $^{*}$: the paired 95\% bootstrap confidence interval excludes zero.}
\label{tab:benchmarks}
\small
\setlength{\tabcolsep}{4pt} 
\resizebox{\textwidth}{!}{%
\begin{tabular}{ll ccc cccc c c}
\toprule
& & \multicolumn{3}{c}{\textit{Broad-coverage exams}} & \multicolumn{4}{c}{\textit{Science}} & \textit{Control} & \\
\cmidrule(lr){3-5} \cmidrule(lr){6-9} \cmidrule(lr){10-10}
\textbf{Model} & & \textbf{MMLU} & \textbf{CMMLU} & \textbf{C-Eval} & \textbf{ChemBench} & \textbf{ScienceQA} & \textbf{SciEval} & \textbf{MoleculeQA} & \textbf{MedMCQA} & \textbf{Mean} \\
\midrule
\multirow{3}{*}{Llama-3.2-3B} & Base & 56.5 & 44.3 & 44.3 & 49.2 & 76.8 & 53.4 & 27.4 & 50.4 & 50.3 \\
 & + WoW & 55.2 & 43.2 & 41.7 & 50.3 & 78.9 & 58.8 & 28.8 & 48.1 & 50.6 \\
 & $\Delta$ & $-1.2^{*}$ & $-1.1^{*}$ & $-2.6^{*}$ & $+1.1$ & $\mathbf{+2.1}^{*}$ & $\mathbf{+5.5}^{*}$ & $\mathbf{+1.4}^{*}$ & $-2.3^{*}$ & $+0.3$ \\
\midrule
\multirow{3}{*}{Llama-3.1-8B} & Base & 65.3 & 52.9 & 52.5 & 53.9 & 82.7 & 64.6 & 29.3 & 58.0 & 57.4 \\
 & + WoW & 65.0 & 52.2 & 52.7 & 54.2 & 82.9 & 65.1 & 31.0 & 57.4 & 57.6 \\
 & $\Delta$ & $-0.3$ & $-0.7$ & $+0.2$ & $+0.3$ & $+0.2$ & $+0.5$ & $\mathbf{+1.7}$ & $-0.6$ & $+0.2$ \\
\midrule
\multirow{3}{*}{Qwen3-4B-Base} & Base & 73.0 & 77.6 & 78.6 & 57.2 & 86.8 & 77.6 & 34.0 & 58.9 & 68.0 \\
 & + WoW & 73.4 & 77.2 & 78.4 & 57.8 & 87.0 & 78.0 & 34.4 & 58.3 & 68.1 \\
 & $\Delta$ & $+0.4$ & $-0.4$ & $-0.2$ & $+0.6$ & $+0.2$ & $+0.4$ & $+0.4$ & $-0.6$ & $+0.1$ \\
\midrule
\multicolumn{2}{l}{\#Items} & 14,042 & 11,582 & 1,346 & 2,147 & 2,224 & 10,704 & 5,786 & 4,183 & -- \\
\bottomrule
\end{tabular}%
}
\end{table*}

Table~\ref{tab:benchmarks} reports knowledge-oriented accuracy on the external benchmarks. For Llama-3.2-3B, fine-tuning improves three of the four science benchmarks significantly: SciEval by $+5.5$ points, ScienceQA by $+2.1$, and MoleculeQA by $+1.4$. Within SciEval, the gain comes from the disciplines our sources cover, with chemistry improving by $+7.3$, biology by $+6.4$, and physics by $+3.9$. Its medical portion (MedQA), which our sources do not cover, decreases by $3.7$ (\S\ref{sec:data}). For the two stronger models, the science benchmarks are stable or improve slightly. Llama-3.1-8B gains $+1.7$ points on MoleculeQA and $+0.5$ on SciEval, and Qwen3-4B-Base gains $+0.6$ on ChemBench and $+0.4$ on both SciEval and MoleculeQA.

Knowledge outside the scope of WoW is largely preserved, and the cost that remains is concentrated in the smallest model. On the broad-coverage exams, about half of whose items lie in subjects our sources do not cover, such as psychology and clinical medicine (Appendix~\ref{app:benchmark-subjects}), Llama-3.2-3B loses 1.1--2.6 points. The two larger models stay within 0.7 points of their base checkpoints, and Qwen3-4B-Base even gains 0.4 points on MMLU. The out-of-domain control shows the same pattern: MedMCQA decreases by 2.3 points for Llama-3.2-3B but by only 0.6 points for Llama-3.1-8B and Qwen3-4B-Base. This is consistent with prior observations that forgetting under continual fine-tuning is more pronounced in smaller models \citep{luo2023empirical}: with more capacity, the new disciplinary supervision displaces less of the existing knowledge.

\section{Conclusion}

We presented WoW, a large-scale educational corpus designed to support full-lifecycle LLM training. Its Educational Knowledge Graph organizes heterogeneous resources under a shared taxonomy, makes knowledge coverage measurable, and guides targeted collection and completion. A four-phase pipeline transforms source documents into grounded, self-contained, and reasoning-intensive supervision while preserving document-level provenance. We release 188K instruction instances together with an interactive platform for graph-based data selection and retrieval.

Fine-tuning three base models on WoW consistently improves disciplinary problem solving and yields modest gains of 0.1--0.3 percentage points in mean accuracy across eight external benchmarks. External gains concentrate on aligned science benchmarks, while the 4B and 8B models largely preserve broad knowledge. These findings support the value of systematically organized educational data for disciplinary specialization. Future work will extend evaluation to open-ended reasoning and the corpus's pre-training and multimodal components.

\bibliography{Reference}

@article{ostling2023cambridge,
  title={The Cambridge law corpus: A dataset for legal AI research},
  author={{\"O}stling, Andreas and Sargeant, Holli and Xie, Huiyuan and Bull, Ludwig and Terenin, Alexander and Jonsson, Leif and Magnusson, M{\aa}ns and Steffek, Felix},
  journal={Advances in Neural Information Processing Systems},
  volume={36},
  pages={41355--41385},
  year={2023}
}

@inproceedings{liu2025cmm,
  title={Cmm-math: A chinese multimodal math dataset to evaluate and enhance the mathematics reasoning of large multimodal models},
  author={Liu, Wentao and Pan, Qianjun and Zhang, Yi and Liu, Zhuo and Wu, Ji and Zhou, Jie and Zhou, Aimin and Chen, Qin and Jiang, Bo and He, Liang},
  booktitle={Proceedings of the 33rd ACM International Conference on Multimedia},
  pages={12585--12591},
  year={2025}
}

@article{penedo2024fineweb,
  title={The fineweb datasets: Decanting the web for the finest text data at scale},
  author={Penedo, Guilherme and Kydl{\'\i}{\v{c}}ek, Hynek and Lozhkov, Anton and Mitchell, Margaret and Raffel, Colin and Von Werra, Leandro and Wolf, Thomas and others},
  journal={Advances in Neural Information Processing Systems},
  volume={37},
  pages={30811--30849},
  year={2024}
}

@article{zhao2022edukg,
  title={EDUKG: a heterogeneous sustainable k-12 educational knowledge graph},
  author={Zhao, Bowen and Sun, Jiuding and Xu, Bin and Lu, Xingyu and Li, Yuchen and Yu, Jifan and Liu, Minghui and Zhang, Tingjian and Chen, Qiuyang and Li, Hanming and others},
  journal={arXiv preprint arXiv:2210.12228},
  year={2022}
}

@misc{gbt13745_2009,
  title        = {{GB/T 13745—2009 Classification and Code of Disciplines}},
  author       = {{Standardization Administration of China}},
  year         = {2009},
  howpublished = {National Standard of the People's Republic of China},
  note         = {Issued on May 6, 2009; implemented on Nov. 1, 2009}
}

@article{osborne2022science,
  title={Science, misinformation, and the role of education},
  author={Osborne, Jonathan and Pimentel, Daniel},
  journal={Science},
  volume={378},
  number={6617},
  pages={246--248},
  year={2022},
  publisher={American Association for the Advancement of Science}
}

@article{badshah2023towards,
  title={Towards smart education through internet of things: A survey},
  author={Badshah, Afzal and Ghani, Anwar and Daud, Ali and Jalal, Ateeqa and Bilal, Muhammad and Crowcroft, Jon},
  journal={ACM Computing Surveys},
  volume={56},
  number={2},
  pages={1--33},
  year={2023},
  publisher={ACM New York, NY}
}

@inproceedings{lo2020s2orc,
  title={S2ORC: The semantic scholar open research corpus},
  author={Lo, Kyle and Wang, Lucy Lu and Neumann, Mark and Kinney, Rodney and Weld, Daniel S},
  booktitle={Proceedings of the 58th annual meeting of the association for computational linguistics},
  pages={4969--4983},
  year={2020}
}

@article{kocetkov2022stack,
  title={The stack: 3 tb of permissively licensed source code},
  author={Kocetkov, Denis and Li, Raymond and Allal, Loubna Ben and Li, Jia and Mou, Chenghao and Ferrandis, Carlos Mu{\~n}oz and Jernite, Yacine and Mitchell, Margaret and Hughes, Sean and Wolf, Thomas and others},
  journal={arXiv preprint arXiv:2211.15533},
  year={2022}
}

@article{henderson2022pile,
  title={Pile of law: Learning responsible data filtering from the law and a 256gb open-source legal dataset},
  author={Henderson, Peter and Krass, Mark and Zheng, Lucia and Guha, Neel and Manning, Christopher D and Jurafsky, Dan and Ho, Daniel},
  journal={Advances in Neural Information Processing Systems},
  volume={35},
  pages={29217--29234},
  year={2022}
}

@inproceedings{wang2022super,
  title={Super-naturalinstructions: Generalization via declarative instructions on 1600+ nlp tasks},
  author={Wang, Yizhong and Mishra, Swaroop and Alipoormolabashi, Pegah and Kordi, Yeganeh and Mirzaei, Amirreza and Naik, Atharva and Ashok, Arjun and Dhanasekaran, Arut Selvan and Arunkumar, Anjana and Stap, David and others},
  booktitle={Proceedings of the 2022 conference on empirical methods in natural language processing},
  pages={5085--5109},
  year={2022}
}

@article{laurenccon2022bigscience,
  title={The bigscience roots corpus: A 1.6 tb composite multilingual dataset},
  author={Lauren{\c{c}}on, Hugo and Saulnier, Lucile and Wang, Thomas and Akiki, Christopher and Villanova del Moral, Albert and Le Scao, Teven and Von Werra, Leandro and Mou, Chenghao and Gonz{\'a}lez Ponferrada, Eduardo and Nguyen, Huu and others},
  journal={Advances in Neural Information Processing Systems},
  volume={35},
  pages={31809--31826},
  year={2022}
}

@inproceedings{niklaus2024multilegalpile,
  title={Multilegalpile: A 689gb multilingual legal corpus},
  author={Niklaus, Joel and Matoshi, Veton and St{\"u}rmer, Matthias and Chalkidis, Ilias and Ho, Daniel},
  booktitle={Proceedings of the 62nd Annual Meeting of the Association for Computational Linguistics (Volume 1: Long Papers)},
  pages={15077--15094},
  year={2024}
}

@inproceedings{soldaini2024dolma,
  title={Dolma: An open corpus of three trillion tokens for language model pretraining research},
  author={Soldaini, Luca and Kinney, Rodney and Bhagia, Akshita and Schwenk, Dustin and Atkinson, David and Authur, Russell and Bogin, Ben and Chandu, Khyathi and Dumas, Jennifer and Elazar, Yanai and others},
  booktitle={Proceedings of the 62nd Annual Meeting of the Association for Computational Linguistics (Volume 1: Long Papers)},
  pages={15725--15788},
  year={2024}
}

@article{li2024datacomp,
  title={Datacomp-lm: In search of the next generation of training sets for language models},
  author={Li, Jeffrey and Fang, Alex and Smyrnis, Georgios and Ivgi, Maor and Jordan, Matt and Gadre, Samir and Bansal, Hritik and Guha, Etash and Keh, Sedrick and Arora, Kushal and others},
  journal={Advances in Neural Information Processing Systems},
  volume={37},
  pages={14200--14282},
  year={2024}
}

@article{han2024infimm,
  title={Infimm-webmath-40b: Advancing multimodal pre-training for enhanced mathematical reasoning},
  author={Han, Xiaotian and Jian, Yiren and Hu, Xuefeng and Liu, Haogeng and Wang, Yiqi and Fan, Qihang and Ai, Yuang and Huang, Huaibo and He, Ran and Yang, Zhenheng and others},
  journal={arXiv preprint arXiv:2409.12568},
  volume={4},
  year={2024}
}

@article{dan2023educhat,
  title={EduChat: A Large-Scale Language Model-based Chatbot System for Intelligent Education},
  author={Dan, Yuhao and Lei, Zhikai and Gu, Yiyang and Li, Yong and Yin, Jianghao and Lin, Jiaju and others},
  journal={arXiv preprint arXiv:2308.02773},
  year={2023},
  url={https://arxiv.org/abs/2308.02773}
}

@article{yu2025opencsg,
  title={OpenCSG Chinese Corpus: A Series of High-quality Chinese Datasets for LLM Training},
  author={Yu, Yijiong and Dai, Ziyun and Wang, Zekun and Wang, Wei and Chen, Ran and Pei, Ji},
  journal={arXiv preprint arXiv:2501.08197},
  year={2025},
  url={https://arxiv.org/abs/2501.08197}
}

@inproceedings{wang2024book2dial,
  title={Book2Dial: Generating Teacher-Student Interactions from Textbooks for Cost-Effective Development of Educational Chatbots},
  author={Wang, Junling and Macina, Jakub and Daheim, Nico and Chowdhury, Sankalan Pal and Sachan, Mrinmaya},
  booktitle={Findings of the Association for Computational Linguistics: ACL 2024},
  year={2024},
  url={https://arxiv.org/abs/2403.03307}
}

@inproceedings{hawasly2025arabicweb,
  title={ArabicWeb-Edu: Educational Quality Data for Arabic LLM Training},
  author={Hawasly, Majd and Mohiuddin, Muhammad Tasnim and Mubarak, Hamdy and Boughorbel, Sabri},
  booktitle={Proceedings of The Third Arabic Natural Language Processing Conference},
  pages={436--447},
  year={2025}
}

@inproceedings{touchent2026biomed,
  title={Biomed-Enriched: Data-Efficient Biomedical Pretraining via Paragraph-Level Annotation},
  author={Touchent, Rian and Godey, Nathan and Clergerie, {\'E}ric},
  booktitle={Findings of the Association for Computational Linguistics: ACL 2026},
  pages={34276--34287},
  year={2026}
}

@inproceedings{lozano2025biomedica,
  title={Biomedica: An open biomedical image-caption archive, dataset, and vision-language models derived from scientific literature},
  author={Lozano, Alejandro and Sun, Min Woo and Burgess, James and Chen, Liangyu and Nirschl, Jeffrey J and Gu, Jeffrey and Lopez, Ivan and Aklilu, Josiah and Rau, Anita and Katzer, Austin Wolfgang and others},
  booktitle={Proceedings of the Computer Vision and Pattern Recognition Conference},
  pages={19724--19735},
  year={2025}
}

@inproceedings{lai2017race,
  title={Race: Large-scale reading comprehension dataset from examinations},
  author={Lai, Guokun and Xie, Qizhe and Liu, Hanxiao and Yang, Yiming and Hovy, Eduard},
  booktitle={Proceedings of the 2017 conference on empirical methods in natural language processing},
  pages={785--794},
  year={2017}
}

@inproceedings{hardalov2020exams,
  title={EXAMS: A Multi-subject High School Examinations Dataset for Cross-lingual and Multilingual Question Answering},
  author={Hardalov, Momchil and Mihaylov, Todor and Zlatkova, Dimitrina and Dinkov, Yoan and Koychev, Ivan and Nakov, Preslav},
  booktitle={Proceedings of the 2020 Conference on Empirical Methods in Natural Language Processing},
  year={2020},
  url={https://aclanthology.org/2020.emnlp-main.438/}
}

@inproceedings{forbes2020social,
  title={Social chemistry 101: Learning to reason about social and moral norms},
  author={Forbes, Maxwell and Hwang, Jena D and Shwartz, Vered and Sap, Maarten and Choi, Yejin},
  booktitle={Proceedings of the 2020 Conference on Empirical Methods in Natural Language Processing (EMNLP)},
  pages={653--670},
  year={2020}
}

@article{hendrycks2020measuring,
  title={Measuring massive multitask language understanding},
  author={Hendrycks, Dan and Burns, Collin and Basart, Steven and Zou, Andy and Mazeika, Mantas and Song, Dawn and Steinhardt, Jacob},
  journal={arXiv preprint arXiv:2009.03300},
  year={2020}
}

@article{hendrycks2020aligning,
  title={Aligning ai with shared human values},
  author={Hendrycks, Dan and Burns, Collin and Basart, Steven and Critch, Andrew and Li, Jerry and Song, Dawn and Steinhardt, Jacob},
  journal={arXiv preprint arXiv:2008.02275},
  year={2020}
}

@inproceedings{emelin2021moral,
  title={Moral stories: Situated reasoning about norms, intents, actions, and their consequences},
  author={Emelin, Denis and Le Bras, Ronan and Hwang, Jena D and Forbes, Maxwell and Choi, Yejin},
  booktitle={Proceedings of the 2021 Conference on Empirical Methods in Natural Language Processing},
  pages={698--718},
  year={2021}
}

@inproceedings{dasigi2021dataset,
  title={A dataset of information-seeking questions and answers anchored in research papers},
  author={Dasigi, Pradeep and Lo, Kyle and Beltagy, Iz and Cohan, Arman and Smith, Noah A and Gardner, Matt},
  booktitle={Proceedings of the 2021 Conference of the North American Chapter of the Association for Computational Linguistics: Human Language Technologies},
  pages={4599--4610},
  year={2021}
}

@inproceedings{zhu2021tat,
  title={TAT-QA: A question answering benchmark on a hybrid of tabular and textual content in finance},
  author={Zhu, Fengbin and Lei, Wenqiang and Huang, Youcheng and Wang, Chao and Zhang, Shuo and Lv, Jiancheng and Feng, Fuli and Chua, Tat-Seng},
  booktitle={Proceedings of the 59th annual meeting of the Association for Computational Linguistics and the 11th international joint conference on natural language processing (volume 1: long papers)},
  pages={3277--3287},
  year={2021}
}

@inproceedings{chen2021finqa,
  title={Finqa: A dataset of numerical reasoning over financial data},
  author={Chen, Zhiyu and Chen, Wenhu and Smiley, Charese and Shah, Sameena and Borova, Iana and Langdon, Dylan and Moussa, Reema and Beane, Matt and Huang, Ting-Hao and Routledge, Bryan R and others},
  booktitle={Proceedings of the 2021 Conference on Empirical Methods in Natural Language Processing},
  pages={3697--3711},
  year={2021}
}

@inproceedings{lu2022scienceqa,
  title={Learn to Explain: Multimodal Reasoning via Thought Chains for Science Question Answering},
  author={Lu, Pan and Mishra, Swaroop and Xia, Tony and Qiu, Liang and Chang, Kai-Wei and Zhu, Song-Chun and Tafjord, Oyvind and Clark, Peter and Kalyan, Ashwin},
  booktitle={Advances in Neural Information Processing Systems},
  year={2022},
  url={https://arxiv.org/abs/2209.09513}
}

@inproceedings{masry2022chartqa,
  title={Chartqa: A benchmark for question answering about charts with visual and logical reasoning},
  author={Masry, Ahmed and Tan, Jia Qing and Joty, Shafiq and Hoque, Enamul and others},
  booktitle={Findings of the association for computational linguistics: ACL 2022},
  pages={2263--2279},
  year={2022}
}

@inproceedings{pal2022medmcqa,
  title={Medmcqa: A large-scale multi-subject multi-choice dataset for medical domain question answering},
  author={Pal, Ankit and Umapathi, Logesh Kumar and Sankarasubbu, Malaikannan},
  booktitle={Conference on health, inference, and learning},
  pages={248--260},
  year={2022},
  organization={PMLR}
}

@inproceedings{NEURIPS2023_6a386d70,
  author = {Xie, Qianqian and Han, Weiguang and Zhang, Xiao and Lai, Yanzhao and Peng, Min and Lopez-Lira, Alejandro and Huang, Jimin},
  booktitle = {Advances in Neural Information Processing Systems},
  editor = {A. Oh and T. Naumann and A. Globerson and K. Saenko and M. Hardt and S. Levine},
  pages = {33469--33484},
  publisher = {Curran Associates, Inc.},
  title = {PIXIU: A Comprehensive Benchmark, Instruction Dataset and Large Language Model for Finance},
  url = {https://proceedings.neurips.cc/paper_files/paper/2023/file/6a386d703b50f1cf1f61ab02a15967bb-Paper-Datasets_and_Benchmarks.pdf},
  volume = {36},
  year = {2023}
}

@inproceedings{huang2023ceval,
  title={C-Eval: A Multi-Level Multi-Discipline Chinese Evaluation Suite for Foundation Models},
  author={Huang, Yuzhen and Bai, Yuzhuo and Zhu, Zhihao and Zhang, Junlei and Zhang, Jinghan and Su, Tangjun and Liu, Junteng and Lv, Chuancheng and Zhang, Yikai and Lei, Jiayi and Fu, Yao and Sun, Maosong and He, Junxian},
  booktitle={Advances in Neural Information Processing Systems},
  year={2023},
  url={https://arxiv.org/abs/2305.08322}
}

@article{li2023cmmlu,
  title={CMMLU: Measuring Massive Multitask Language Understanding in Chinese},
  author={Li, Haonan and Zhang, Yixuan and Koto, Fajri and Yang, Yifei and Zhao, Hai and Gong, Yeyun and Duan, Nan and Baldwin, Timothy},
  journal={arXiv preprint arXiv:2306.09212},
  year={2023},
  url={https://arxiv.org/abs/2306.09212}
}

@article{liu2023m3ke,
  title={M3KE: A Massive Multi-Level Multi-Subject Knowledge Evaluation Benchmark for Chinese Large Language Models},
  author={Liu, Chuang and Jin, Renren and Ren, Yuxi and Wu, Yujie and Liu, Dong and Liu, Ji and Liu, Xiaokang and Liu, Deyi},
  journal={arXiv preprint arXiv:2305.10263},
  year={2023},
  url={https://arxiv.org/abs/2305.10263}
}

@article{zhang2023m3exam,
  title={M3Exam: A Multilingual, Multimodal, Multilevel Benchmark for Examining Large Language Models},
  author={Zhang, Wenxuan and Aljunied, Sharifah Mahani and Gao, Chang and Chia, Yew Ken and Bing, Lidong},
  journal={arXiv preprint arXiv:2306.05179},
  year={2023},
  url={https://arxiv.org/abs/2306.05179}
}

@article{guha2023legalbench,
  title={Legalbench: A collaboratively built benchmark for measuring legal reasoning in large language models},
  author={Guha, Neel and Nyarko, Julian and Ho, Daniel and R{\'e}, Christopher and Chilton, Adam and Chohlas-Wood, Alex and Peters, Austin and Waldon, Brandon and Rockmore, Daniel and Zambrano, Diego and others},
  journal={Advances in neural information processing systems},
  volume={36},
  pages={44123--44279},
  year={2023}
}

@article{hou2024eeval,
  title={E-EVAL: A Comprehensive Chinese K-12 Education Evaluation Benchmark for Large Language Models},
  author={Hou, Jinchang and Ao, Chang and Wu, Haihong and Kong, Xiangtao and Zheng, Zhigang and Tang, Daijia and Li, Chengming and Hu, Xiping and Xu, Ruifeng and Ni, Shiwen and Yang, Min},
  journal={arXiv preprint arXiv:2401.15927},
  year={2024},
  url={https://arxiv.org/abs/2401.15927}
}

@inproceedings{you2024ck12,
  title={CK12: a rounded K12 knowledge graph based benchmark for Chinese holistic cognition evaluation},
  author={You, Weihao and Wang, Pengcheng and Li, Changlong and Ji, Zhilong and Bai, Jinfeng},
  booktitle={Proceedings of the AAAI Conference on Artificial Intelligence},
  volume={38},
  number={17},
  pages={19431--19439},
  year={2024}
}

@inproceedings{yue2024mmmu,
  title={Mmmu: A massive multi-discipline multimodal understanding and reasoning benchmark for expert agi},
  author={Yue, Xiang and Ni, Yuansheng and Zhang, Kai and Zheng, Tianyu and Liu, Ruoqi and Zhang, Ge and Stevens, Samuel and Jiang, Dongfu and Ren, Weiming and Sun, Yuxuan and others},
  booktitle={Proceedings of the IEEE/CVF conference on computer vision and pattern recognition},
  pages={9556--9567},
  year={2024}
}

@article{zhang2024cmmmu,
  title={CMMMU: A Chinese Massive Multi-discipline Multimodal Understanding Benchmark},
  author={Zhang, Ge and Du, Xinrun and Chen, Bei and Liang, Yiming and Luo, Tongxu and others},
  journal={arXiv preprint arXiv:2401.11944},
  year={2024},
  url={https://arxiv.org/abs/2401.11944}
}

@article{das2024examsv,
  title={EXAMS-V: A Multi-Discipline Multilingual Multimodal Exam Benchmark for Evaluating Vision Language Models},
  author={Das, Rocktim Jyoti and Hristov, Simeon Emilov and Li, Haonan and Dimitrov, Dimitar Iliyanov and Koychev, Ivan and Nakov, Preslav},
  journal={arXiv preprint arXiv:2403.10378},
  year={2024},
  url={https://arxiv.org/abs/2403.10378}
}

@inproceedings{ren2024ceamc,
  title={Ceamc: corpus and empirical study of argument analysis in education via LLMs},
  author={Ren, Yupei and Wu, Hongyi and Long, Zhaoguang and Zhao, Shangqing and Zhou, Xinyi and Yin, Zheqin and Zhuang, Xinlin and Bai, Xiaopeng and Lan, Man},
  booktitle={Findings of the Association for Computational Linguistics: EMNLP 2024},
  pages={6949--6966},
  year={2024}
}

@inproceedings{song2023matsci,
  title={MatSci-NLP: Evaluating scientific language models on materials science language tasks using text-to-schema modeling},
  author={Song, Yu and Miret, Santiago and Liu, Bang},
  booktitle={Proceedings of the 61st Annual Meeting of the Association for Computational Linguistics (Volume 1: Long Papers)},
  pages={3621--3639},
  year={2023}
}

@inproceedings{jimenez2024swe,
  title={Swe-bench: Can language models resolve real-world github issues?},
  author={Jimenez, Carlos E and Yang, John and Wettig, Alexander and Yao, Shunyu and Pei, Kexin and Press, Ofir and Narasimhan, Karthik},
  booktitle={International Conference on Learning Representations},
  volume={2024},
  pages={54107--54157},
  year={2024}
}

@article{gunasekar2023textbooks,
  title   = {Textbooks Are All You Need},
  author  = {Gunasekar, Suriya and Zhang, Yi and Aneja, Jyoti and Mendes, Caio C{\'e}sar Teodoro and Del Giorno, Allie and Gopi, Sivakanth and Javaheripi, Mojan and Kauffmann, Piero and de Rosa, Gustavo and Saarikivi, Olli and others},
  journal = {arXiv preprint arXiv:2306.11644},
  year    = {2023}
}

@article{li2023textbooks,
  title   = {Textbooks Are All You Need {II}: {phi-1.5} Technical Report},
  author  = {Li, Yuanzhi and Bubeck, S{\'e}bastien and Eldan, Ronen and Del Giorno, Allie and Gunasekar, Suriya and Lee, Yin Tat},
  journal = {arXiv preprint arXiv:2309.05463},
  year    = {2023}
}

@inproceedings{zhou2023lima,
  title     = {{LIMA}: Less Is More for Alignment},
  author    = {Zhou, Chunting and Liu, Pengfei and Xu, Puxin and Iyer, Srini and Sun, Jiao and Mao, Yuning and Ma, Xuezhe and Efrat, Avia and Yu, Ping and Yu, Lili and others},
  booktitle = {Advances in Neural Information Processing Systems},
  volume    = {36},
  year      = {2023}
}

@inproceedings{paster2024openwebmath,
  title     = {{OpenWebMath}: An Open Dataset of High-Quality Mathematical Web Text},
  author    = {Paster, Keiran and Dos Santos, Marco and Azerbayev, Zhangir and Ba, Jimmy},
  booktitle = {The Twelfth International Conference on Learning Representations},
  year      = {2024}
}

@article{shao2024deepseekmath,
  title   = {{DeepSeekMath}: Pushing the Limits of Mathematical Reasoning in Open Language Models},
  author  = {Shao, Zhihong and Wang, Peiyi and Zhu, Qihao and Xu, Runxin and Song, Junxiao and others},
  journal = {arXiv preprint arXiv:2402.03300},
  year    = {2024}
}

@inproceedings{azerbayev2024llemma,
  title     = {Llemma: An Open Language Model for Mathematics},
  author    = {Azerbayev, Zhangir and Schoelkopf, Hailey and Paster, Keiran and Dos Santos, Marco and McAleer, Stephen and Jiang, Albert Q. and Deng, Jia and Biderman, Stella and Welleck, Sean},
  booktitle = {The Twelfth International Conference on Learning Representations},
  year      = {2024}
}

@inproceedings{longpre2023flan,
  title     = {The Flan Collection: Designing Data and Methods for Effective Instruction Tuning},
  author    = {Longpre, Shayne and Hou, Le and Vu, Tu and Webson, Albert and Chung, Hyung Won and Tay, Yi and Zhou, Denny and Le, Quoc V. and Zoph, Barret and Wei, Jason and Roberts, Adam},
  booktitle = {Proceedings of the 40th International Conference on Machine Learning},
  series    = {Proceedings of Machine Learning Research},
  volume    = {202},
  year      = {2023}
}

@inproceedings{wang2023selfinstruct,
  title     = {Self-Instruct: Aligning Language Models with Self-Generated Instructions},
  author    = {Wang, Yizhong and Kordi, Yeganeh and Mishra, Swaroop and Liu, Alisa and Smith, Noah A. and Khashabi, Daniel and Hajishirzi, Hannaneh},
  booktitle = {Proceedings of the 61st Annual Meeting of the Association for Computational Linguistics (Volume 1: Long Papers)},
  pages     = {13484--13508},
  year      = {2023}
}

@inproceedings{xu2024wizardlm,
  title     = {{WizardLM}: Empowering Large Pre-Trained Language Models to Follow Complex Instructions},
  author    = {Xu, Can and Sun, Qingfeng and Zheng, Kai and Geng, Xiubo and Zhao, Pu and Feng, Jiazhan and Tao, Chongyang and Lin, Qingwei and Jiang, Daxin},
  booktitle = {The Twelfth International Conference on Learning Representations},
  year      = {2024}
}

@article{mukherjee2023orca,
  title   = {Orca: Progressive Learning from Complex Explanation Traces of {GPT-4}},
  author  = {Mukherjee, Subhabrata and Mitra, Arindam and Jawahar, Ganesh and Agarwal, Sahaj and Palangi, Hamid and Awadallah, Ahmed},
  journal = {arXiv preprint arXiv:2306.02707},
  year    = {2023}
}

@inproceedings{cheng2024adapting,
  title     = {Adapting Large Language Models via Reading Comprehension},
  author    = {Cheng, Daixuan and Huang, Shaohan and Wei, Furu},
  booktitle = {The Twelfth International Conference on Learning Representations},
  year      = {2024}
}

@inproceedings{cheng2024instruction,
  title     = {Instruction Pre-Training: Language Models are Supervised Multitask Learners},
  author    = {Cheng, Daixuan and Gu, Yuxian and Huang, Shaohan and Bi, Junyu and Huang, Minlie and Wei, Furu},
  booktitle = {Proceedings of the 2024 Conference on Empirical Methods in Natural Language Processing},
  year      = {2024}
}

@inproceedings{yue2024mammoth2,
  title     = {{MAmmoTH2}: Scaling Instructions from the Web},
  author    = {Yue, Xiang and Zheng, Tuney and Zhang, Ge and Chen, Wenhu},
  booktitle = {Advances in Neural Information Processing Systems},
  volume    = {37},
  year      = {2024}
}

@article{gudibande2023false,
  title   = {The False Promise of Imitating Proprietary {LLMs}},
  author  = {Gudibande, Arnav and Wallace, Eric and Snell, Charlie and Geng, Xinyang and Liu, Hao and Abbeel, Pieter and Levine, Sergey and Song, Dawn},
  journal = {arXiv preprint arXiv:2305.15717},
  year    = {2023}
}

@inproceedings{wei2022chain,
  title     = {Chain-of-Thought Prompting Elicits Reasoning in Large Language Models},
  author    = {Wei, Jason and Wang, Xuezhi and Schuurmans, Dale and Bosma, Maarten and Ichter, Brian and Xia, Fei and Chi, Ed and Le, Quoc V. and Zhou, Denny},
  booktitle = {Advances in Neural Information Processing Systems},
  volume    = {35},
  pages     = {24824--24837},
  year      = {2022}
}

@article{luo2023empirical,
  title   = {An Empirical Study of Catastrophic Forgetting in Large Language Models During Continual Fine-tuning},
  author  = {Luo, Yun and Yang, Zhen and Meng, Fandong and Li, Yafu and Zhou, Jie and Zhang, Yue},
  journal = {arXiv preprint arXiv:2308.08747},
  year    = {2023}
}

@article{grattafiori2024llama3,
  title   = {The {Llama} 3 Herd of Models},
  author  = {Grattafiori, Aaron and Dubey, Abhimanyu and Jauhri, Abhinav and Pandey, Abhinav and Kadian, Abhishek and others},
  journal = {arXiv preprint arXiv:2407.21783},
  year    = {2024}
}

@article{yang2025qwen3,
  title   = {{Qwen3} Technical Report},
  author  = {Yang, An and Li, Anfeng and Yang, Baosong and Zhang, Beichen and Hui, Binyuan and others},
  journal = {arXiv preprint arXiv:2505.09388},
  year    = {2025}
}

@software{gao2024lmeval,
  title     = {A Framework for Few-Shot Language Model Evaluation},
  author    = {Gao, Leo and Tow, Jonathan and Abbasi, Baber and Biderman, Stella and Black, Sid and others},
  publisher = {Zenodo},
  doi       = {10.5281/zenodo.10256836},
  year      = {2024}
}

@inproceedings{sun2024scieval,
  title={Scieval: A multi-level large language model evaluation benchmark for scientific research},
  author={Sun, Liangtai and Han, Yang and Zhao, Zihan and Ma, Da and Shen, Zhennan and Chen, Baocai and Chen, Lu and Yu, Kai},
  booktitle={Proceedings of the AAAI Conference on Artificial Intelligence},
  volume={38},
  number={17},
  pages={19053--19061},
  year={2024}
}

@article{wang2023scibench,
  title={Scibench: Evaluating college-level scientific problem-solving abilities of large language models},
  author={Wang, Xiaoxuan and Hu, Ziniu and Lu, Pan and Zhu, Yanqiao and Zhang, Jieyu and Subramaniam, Satyen and Loomba, Arjun R and Zhang, Shichang and Sun, Yizhou and Wang, Wei},
  journal={arXiv preprint arXiv:2307.10635},
  year={2023}
}

@article{zhang2024sciinstruct,
  title={Sciinstruct: a self-reflective instruction annotated dataset for training scientific language models},
  author={Zhang, Dan and Hu, Ziniu and Zhoubian, Sining and Du, Zhengxiao and Yang, Kaiyu and Wang, Zihan and Yue, Yisong and Dong, Yuxiao and Tang, Jie},
  journal={Advances in Neural Information Processing Systems},
  volume={37},
  pages={1443--1473},
  year={2024}
}

@inproceedings{fang2024mol,
  title={Mol-instructions: A large-scale biomolecular instruction dataset for large language models},
  author={Fang, Yin and Liang, Xiaozhuan and Zhang, Ningyu and Liu, Kangwei and Huang, Rui and Chen, Zhuo and Fan, Xiaohui and Chen, Huajun},
  booktitle={International Conference on Learning Representations},
  volume={2024},
  pages={48221--48251},
  year={2024}
}

@article{yu2024llasmol,
  title={Llasmol: Advancing large language models for chemistry with a large-scale, comprehensive, high-quality instruction tuning dataset},
  author={Yu, Botao and Baker, Frazier N and Chen, Ziqi and Ning, Xia and Sun, Huan},
  journal={arXiv preprint arXiv:2402.09391},
  year={2024}
}

@article{mirza2026chempile,
  title={ChemPile: A 250 GB Diverse and Curated Dataset for Chemical Foundation Models},
  author={Mirza, Adrian and Alampara, Nawaf and R{\'\i}os-Garc{\'\i}a, Marti{\~n}o and Abdelalim, Mohamed and Butler, Jack and Connolly, Bethany and Dogan, Tunca and Nezhurina, Marianna and {\c{S}}en, B{\"u}nyamin and Tirunagari, Santosh and others},
  journal={Advances in Neural Information Processing Systems},
  volume={38},
  year={2026}
}

@article{liang2026k12,
  title={K12-KGraph: A Curriculum-Aligned Knowledge Graph for Benchmarking and Training Educational LLMs},
  author={Liang, Hao and Lin, Qihan and Han, Zhaoyang and Ma, Xiaochen and Wong, Zhen Hao and Qiang, Meiyi and Sun, Linzhuang and Zhang, Wentao},
  journal={arXiv preprint arXiv:2605.09635},
  year={2026}
}

@article{laurent2024lab,
  title={Lab-bench: Measuring capabilities of language models for biology research},
  author={Laurent, Jon M and Janizek, Joseph D and Ruzo, Michael and Hinks, Michaela M and Hammerling, Michael J and Narayanan, Siddharth and Ponnapati, Manvitha and White, Andrew D and Rodriques, Samuel G},
  journal={arXiv preprint arXiv:2407.10362},
  year={2024}
}

@article{huang2024olympicarena,
  title={Olympicarena: Benchmarking multi-discipline cognitive reasoning for superintelligent ai},
  author={Huang, Zhen and Wang, Zengzhi and Xia, Shijie and Li, Xuefeng and Zou, Haoyang and Xu, Ruijie and Fan, Run-Ze and Ye, Lyumanshan and Chern, Ethan and Ye, Yixin and others},
  journal={Advances in Neural Information Processing Systems},
  volume={37},
  pages={19209--19253},
  year={2024}
}

@inproceedings{zhou2026mdk12,
  title={Mdk12-bench: a multi-discipline benchmark for evaluating reasoning in multimodal large language models},
  author={Zhou, Pengfei and Peng, Xiaopeng and Zhang, Fanrui and Xu, Zhaopan and Ai, Jiaxin and Qiu, Yansheng and Zhao, Wangbo and Song, Jiajun and Li, Chuanhao and Tang, Weidong and others},
  booktitle={Proceedings of the AAAI Conference on Artificial Intelligence},
  volume={40},
  number={34},
  pages={28982--28990},
  year={2026}
}

@inproceedings{zhang2025physreason,
  title={Physreason: A comprehensive benchmark towards physics-based reasoning},
  author={Zhang, Xinyu and Dong, Yuxuan and Wu, Yanrui and Huang, Jiaxing and Jia, Chengyou and Fernando, Basura and Shou, Mike Zheng and Zhang, Lingling and Liu, Jun},
  booktitle={Proceedings of the 63rd Annual Meeting of the Association for Computational Linguistics (Volume 1: Long Papers)},
  pages={16593--16615},
  year={2025}
}

@article{xiang2026seephys,
  title={Seephys: Does seeing help thinking?--benchmarking vision-based physics reasoning},
  author={Xiang, Kun and Li, Heng and Zhang, Terry Jingchen and Huang, Yinya and Liu, Zirong and Qu, Peixin and He, Jixi and Chen, Jiaqi and Yuan, Yu-Jie and Han, Jianhua and others},
  journal={Advances in Neural Information Processing Systems},
  volume={38},
  year={2026}
}

@article{xu2025ugphysics,
  title={Ugphysics: A comprehensive benchmark for undergraduate physics reasoning with large language models},
  author={Xu, Xin and Xu, Qiyun and Xiao, Tong and Chen, Tianhao and Yan, Yuchen and Zhang, Jiaxin and Diao, Shizhe and Yang, Can and Wang, Yang},
  journal={arXiv preprint arXiv:2502.00334},
  year={2025}
}

@article{wang2025physunibench,
  title={PhysUniBench: A Multi-Modal Physics Reasoning Benchmark at Undergraduate Level},
  author={Wang, Lintao and Su, Encheng and Liu, Jiaqi and Li, Pengze and Xiao, Jiabei and Zhang, Wenlong and Chen, Xi and Meng, Yuan and BAI, LEI and Ouyang, Wanli and others},
  year={2025}
}

@inproceedings{he2024olympiadbench,
  title={Olympiadbench: A challenging benchmark for promoting agi with olympiad-level bilingual multimodal scientific problems},
  author={He, Chaoqun and Luo, Renjie and Bai, Yuzhuo and Hu, Shengding and Thai, Zhen and Shen, Junhao and Hu, Jinyi and Han, Xu and Huang, Yujie and Zhang, Yuxiang and others},
  booktitle={Proceedings of the 62nd Annual Meeting of the Association for Computational Linguistics (Volume 1: Long Papers)},
  pages={3828--3850},
  year={2024}
}

@inproceedings{lu2024moleculeqa,
  title={Moleculeqa: A dataset to evaluate factual accuracy in molecular comprehension},
  author={Lu, Xingyu and Cao, He and Liu, Zijing and Bai, Shengyuan and Chen, Leqing and Yao, Yuan and Zheng, Hai-Tao and Li, Yu},
  booktitle={Findings of the Association for Computational Linguistics: EMNLP 2024},
  pages={3769--3789},
  year={2024}
}

@inproceedings{li2025chemvlm,
  title={Chemvlm: Exploring the power of multimodal large language models in chemistry area},
  author={Li, Junxian and Zhang, Di and Wang, Xunzhi and Hao, Zeying and Lei, Jingdi and Tan, Qian and Zhou, Cai and Liu, Wei and Yang, Yaotian and Xiong, Xinrui and others},
  booktitle={Proceedings of the AAAI Conference on Artificial Intelligence},
  volume={39},
  number={1},
  pages={415--423},
  year={2025}
}

@article{mirza2024large,
  title={Are large language models superhuman chemists?},
  author={Mirza, Adrian and Alampara, Nawaf and Kunchapu, Sreekanth and R{\'\i}os-Garc{\'\i}a, Marti{\~n}o and Emoekabu, Benedict and Krishnan, Aswanth and Gupta, Tanya and Schilling-Wilhelmi, Mara and Okereke, Macjonathan and Aneesh, Anagha and others},
  journal={arXiv preprint arXiv:2404.01475},
  year={2024}
}

\appendix

\setcounter{tocdepth}{2}
\tableofcontents

\section{Experimental Settings}
\label{app:settings}


\paragraph{Models.}
We fine-tune three pre-trained base checkpoints: Llama-3.2-3B, Llama-3.1-8B, and Qwen3-4B-Base. The Llama base checkpoints ship without a chat template; we use the Llama~3 turn format (\texttt{<|start\_header\_id|>}role\texttt{<|end\_header\_id|>} \dots\ \texttt{<|eot\_id|>}) without the date preamble of the instruct template, so that rendering is deterministic. Qwen3-4B-Base uses its own ChatML template with an empty thinking block, i.e.\ the non-thinking format of Qwen3. The template is saved with each fine-tuned checkpoint.

\paragraph{Training data.}
The training split contains 183,048 instances (2,122 for validation and 3,175 for testing), split at the document level with stratification by source collection and question type. Instances longer than 4,096 tokens are dropped rather than truncated (48 instances for the Llama tokenizer, 57 for Qwen), so that no derivation is cut off before its conclusion. We train on the \texttt{messages} schema without its system prompt. The loss covers only the assistant turn, including its end-of-turn token. The selected-response items carry the three format corrections described in \S\ref{sec:exp-setup}: reasoning before the conclusion (44,166 items rewritten), single-answer re-framing of 7,235 items, which makes single-answer items 25.0\% of the multiple-answer pool (previously 1.8\%), and option shuffling, which brings the gold-letter distribution of single-answer items from (A 21.2, B 64.0, C 12.2, D 2.6)\% to (25.0, 27.4, 24.4, 23.2)\%. Options are shuffled only when every reference to an option letter in the answer can be rewritten unambiguously (46,172 of 51,999 items); the remaining items keep their original order.

\paragraph{Hyperparameters.}
Table~\ref{tab:hyperparams} lists the training configuration. All models use the same optimization settings; for Llama-3.1-8B, the per-device batch is halved and gradient accumulation doubled to fit memory, which leaves the effective batch unchanged.

\begin{table}[h]
\centering
\caption{Training configuration.}
\label{tab:hyperparams}
\small
\begin{tabular}{ll}
\toprule
Fine-tuning & full-parameter SFT \\
Epochs & 1 ($\approx$1,430 steps) \\
Effective batch size & 128 sequences \\
Per-device batch $\times$ accumulation & $4 \times 8$ (8B: $2 \times 16$) \\
Max sequence length & 4,096 tokens \\
Optimizer & AdamW ($\beta_1{=}0.9$, $\beta_2{=}0.95$) \\
Peak learning rate & $5\times10^{-6}$ \\
Schedule & cosine, 3\% linear warm-up \\
Weight decay / grad.\ clipping & 0.0 / 1.0 \\
Precision & bf16 compute, fp32 master weights \\
Parallelism & FSDP (full shard) + activation ckpt. \\
Batching & length-grouped \\
Seed & 42 \\
\bottomrule
\end{tabular}
\end{table}

\paragraph{Hardware and software.}
Each model is trained on four NVIDIA A800-SXM4 80\,GB GPUs. Training takes 2.4 hours for Llama-3.2-3B, 3.3 hours for Qwen3-4B-Base, and 5.3 hours for Llama-3.1-8B. Final validation losses are 0.712, 0.553, and 0.583, respectively. We use PyTorch 2.12, Transformers 4.57.6, and Accelerate 1.13 for training, and vLLM 0.11 for inference.

\paragraph{Held-out evaluation.}
Generation is greedy with a limit of 4,096 new tokens. Base and fine-tuned models receive the same plain-text prompt (\texttt{Question: \dots\ Answer:}) preceded by three exemplars of the same question type, sampled with a fixed seed from the training split, and generation stops at the next \texttt{Question:}. The chosen options are extracted with rules that look for an explicit answer statement (e.g.\ ``The correct answer is \textbf{B}'', ``Answer: A, C''). The rules take the last such statement, ignore negated mentions (``the incorrect option is C''), and map statement numbers to the matching option in combination items (``1, 2, and 4''). A prediction is correct if the extracted set equals the gold set; multiple-answer items are additionally scored by F1. Items without an extractable answer are counted as wrong.

\paragraph{External benchmarks.}
All benchmarks are evaluated without a chat template. A prompt ends with the benchmark's answer cue (\texttt{Answer:} or its Chinese equivalent in CMMLU and C-Eval), and the prediction is the option letter with the highest probability among the 20 most likely next tokens. MMLU \citep{hendrycks2020measuring} uses its test split and the five development examples of each subject with the original prompt. CMMLU \citep{li2023cmmlu} uses its test split, and C-Eval \citep{huang2023ceval} its validation split, because test answers are not public; both are 5-shot from their development splits with the answer-only prompts of the original papers. ChemBench \citep{mirza2024large} is restricted to the 2,147 items with a single correct option (2--9 options), since one next-token prediction cannot express an answer set or a number. ScienceQA \citep{lu2022scienceqa} uses the 2,224 test items without an image. SciEval \citep{sun2024scieval} uses the 10,704 multiple-choice items of its locally released test set. MoleculeQA \citep{lu2024moleculeqa} uses its test split with the molecule given as SMILES. MedMCQA \citep{pal2022medmcqa} uses its validation split. All are 0-shot except the three exams. Confidence intervals come from 2,000 paired bootstrap resamples of the per-item differences.

\subsection{Subject Composition of the External Benchmarks}
\label{app:benchmark-subjects}

Table~\ref{tab:exam-subjects} groups the subjects of the three broad-coverage exams into the disciplines of our sources, using keyword rules on the subject names. More than half of the items in each exam fall outside these disciplines; the largest such subjects are moral scenarios, psychology, history, philosophy, and clinical medicine in MMLU, Chinese history, literature, culture, and medicine in CMMLU, and civil-service, teacher-qualification, and physician examinations in C-Eval. Table~\ref{tab:science-subjects} describes the composition of the science benchmarks and the control.

\begin{table}[h]
\centering
\caption{Discipline composition of the broad-coverage exams (number of items and share). \textit{Other} collects subjects outside the disciplines of our sources.}
\label{tab:exam-subjects}
\small
\begin{tabular}{lrrr}
\toprule
\textbf{Discipline} & \textbf{MMLU} & \textbf{CMMLU} & \textbf{C-Eval} \\
\midrule
Mathematics & 1,064 (7.6\%) & 711 (6.1\%) & 90 (6.7\%) \\
Physics & 640 (4.6\%) & 422 (3.6\%) & 57 (4.2\%) \\
Chemistry & 303 (2.2\%) & 132 (1.1\%) & 63 (4.7\%) \\
Biology & 855 (6.1\%) & 662 (5.7\%) & 62 (4.6\%) \\
Computer science & 557 (4.0\%) & 669 (5.8\%) & 133 (9.9\%) \\
Law & 1,763 (12.6\%) & 1,129 (9.7\%) & 47 (3.5\%) \\
Economics \& mgmt. & 1,461 (10.4\%) & 1,072 (9.3\%) & 186 (13.8\%) \\
Other & 7,399 (52.7\%) & 6,785 (58.6\%) & 708 (52.6\%) \\
\midrule
Total & 14,042 & 11,582 & 1,346 \\
\bottomrule
\end{tabular}
\end{table}

\begin{table*}[t]
\centering
\caption{Composition of the science benchmarks and the out-of-domain control, restricted to the items we evaluate.}
\label{tab:science-subjects}
\small
\begin{tabular}{lrp{0.62\textwidth}}
\toprule
\textbf{Benchmark} & \textbf{Items} & \textbf{Composition} \\
\midrule
ChemBench & 2,147 & Single-answer items. Chemical preference 1,000 (pairwise choice between candidate molecules, hence two options), organic 384, toxicity and safety 341, analytical 99, general 98, physical 83, materials science 57, inorganic 49, technical chemistry 36. \\
ScienceQA & 2,224 & Text-only test items (K--12). Natural science 1,043 (biology 397, units and measurement 207, physics 192, chemistry 119, earth science 65), language science 1,056 (writing strategies, figurative language, reference skills, punctuation, grammar), social science 125 (mainly economics and civics). 2--4 options. \\
SciEval & 10,704 & Multiple-choice items of the test set. Chemistry 5,796, biology 2,288, and physics 1,478 (SocraticQA), and medicine 1,142 (MedQA). \\
MoleculeQA & 5,786 & Questions about a molecule given as SMILES, covering its structure, source, properties, and applications; four options. \\
MedMCQA & 4,183 & Validation split of Indian medical entrance exams: dental 1,318, surgery 369, pathology 337, medicine 295, pharmacology 243, anatomy 234, pediatrics 234, and 14 further clinical and basic-science subjects. Our sources contain almost no medicine. \\
\bottomrule
\end{tabular}
\end{table*}

\section{More Implementation Details}
\label{app:implementation}

\subsection{Educational Knowledge Graph Construction}
\label{app:kg}

\begin{figure}[htbp]
    \centering
    \includegraphics[width=\linewidth]{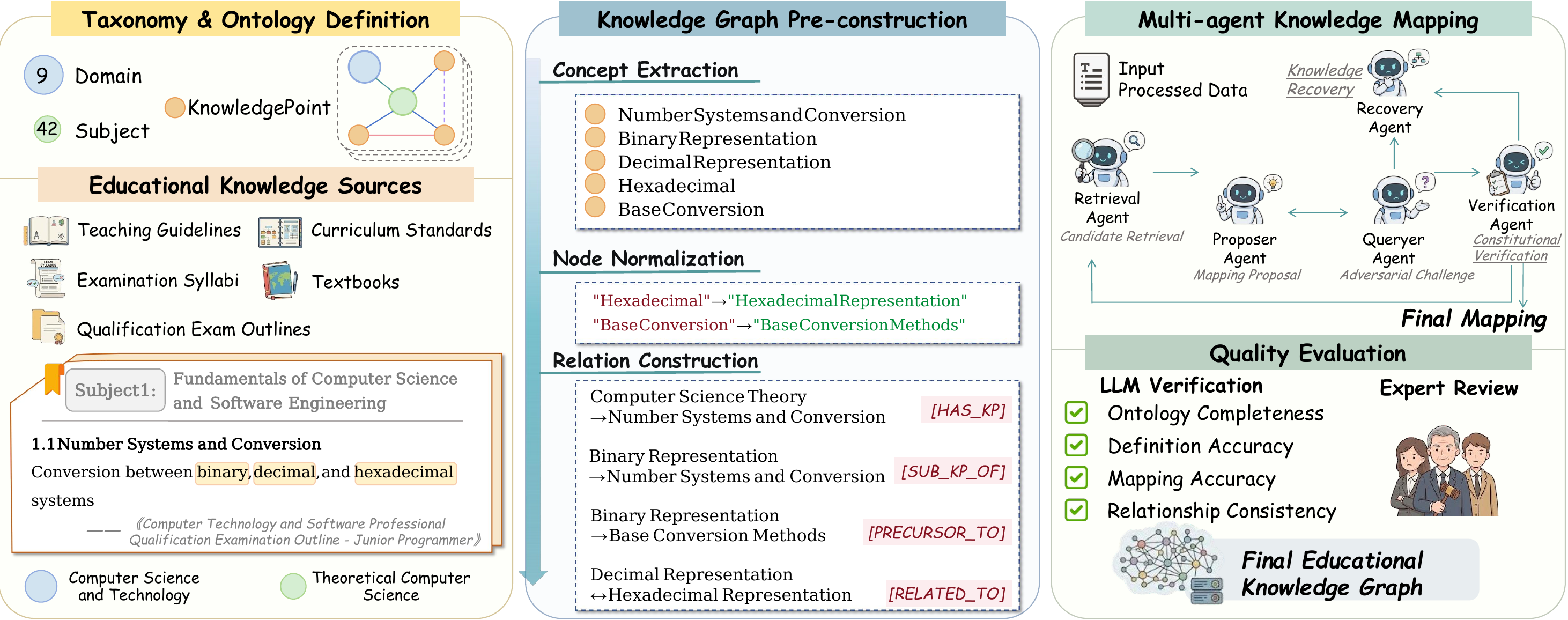}
    \caption{Overview of the Educational Knowledge Graph construction framework}
    \label{fig:kg_construction_framework}
\end{figure}

Existing educational datasets are often organized by individual disciplines and represented with catalog-style or tree-based structures \cite{penedo2024fineweb,liu2025cmm,ostling2023cambridge}. These structures support retrieval within a discipline, but they are less effective in modeling cross disciplinary links, concept aliases, granularity differences, and prerequisite dependencies in educational knowledge \cite{zhao2022edukg}. To address this issue, we construct the Educational Knowledge Graph WoW-EKG as the central structure of the dataset. WoW-EKG is guided by a unified educational ontology and uses KnowledgePoints as the minimal indexable units. It maps educational content from different stages, disciplines, and sources into a shared semantic space. As shown in Figure~\ref{fig:kg_construction_framework}, the construction process includes ontology and taxonomy definition, knowledge graph pre-construction, multi-agent knowledge mapping, and quality evaluation.

\subsection{Ontology Design and Taxonomy}

We define the educational knowledge graph as a structured semantic system consisting of entity types, relation types, and attribute constraints. The ontology contains three entity levels, seven relation types, and corresponding attribute specifications. Entity and relation definitions follow educational cognition principles and national education standards to ensure structural consistency and reliability.

\subsubsection{Entity Hierarchy}
The ontology organizes entities into three levels.

\begin{itemize}
    \item \textbf{Domain.} Domain is the top-level educational category and serves as the entry point for global knowledge indexing. Its design follows official discipline classification standards \cite{gbt13745_2009}. It covers both disciplinary education and literacy oriented education. Disciplinary education includes Chinese, Mathematics, English, Physics, Chemistry, Biology, and History. Literacy oriented education includes Core Literacy and General Competence, which correspond to value formation and key ability development \cite{osborne2022science,badshah2023towards}. Each domain includes an Other category to accommodate emerging interdisciplinary and peripheral knowledge.
    \item \textbf{Subject.} Subject is the second-level category under each domain. Subjects are defined according to official education standards, with fallback categories retained for later expansion.
    \item \textbf{KnowledgePoint.} KnowledgePoint is the core unit of the graph and the minimal indexable unit. Each KnowledgePoint represents an educational concept or skill that can be taught, retrieved, and assessed independently. Each node contains a globally unique ID, a standard name, and a concise academic definition.
\end{itemize}

This hierarchy reduces semantic heterogeneity across sources while preserving flexibility at the KnowledgePoint level. It keeps the graph aligned with official standards and allows adaptation to different education stages and textbook versions.

\subsubsection{Relation Types}

As shown in Table~\ref{tab:relation_schema} Seven semantic relation types are defined according to educational cognition and application needs \cite{zhao2022edukg}. Mandatory relations form the basic graph structure, while recommended and optional relations provide additional semantic links.

\begin{table*}[htbp]
    \centering
    \caption{Education-Oriented Semantic Relation Schema}
    \label{tab:relation_schema}
    \resizebox{\textwidth}{!}{
    \begin{tabular}{cclc}
    \toprule
    Relation Type & Source $\rightarrow$ Target & Semantic Meaning & Level \\
    \midrule
    \texttt{HAS\_SUBJECT} & Domain $\rightarrow$ Subject & A domain contains a subject & Mandatory \\
    \texttt{HAS\_KP} & Subject $\rightarrow$ KnowledgePoint & A subject contains a knowledge point & Mandatory \\
    \texttt{PRECURSOR\_TO} & KnowledgePoint $\rightarrow$ KnowledgePoint & A knowledge point is a prerequisite for another & Mandatory \\
    \texttt{SUB\_KP\_OF} & KnowledgePoint $\rightarrow$ KnowledgePoint & A knowledge point is a subpoint of another & Mandatory \\
    \texttt{RELATED\_TO} & KnowledgePoint $\rightarrow$ KnowledgePoint & Two knowledge points are laterally related & Recommended \\
    \texttt{SAME\_AS} & KnowledgePoint $\rightarrow$ KnowledgePoint & Two knowledge points are equivalent or aliases & Recommended \\
    \texttt{CONTRAST\_TO} & KnowledgePoint $\rightarrow$ KnowledgePoint & Two knowledge points are easily confused and should be contrasted & Optional \\
    \bottomrule
    \end{tabular}}
\end{table*}

\subsection{Knowledge Node Pre-construction}

Knowledge node pre-construction extracts candidate KnowledgePoints from authoritative educational materials and builds initial node and relation sets under ontology constraints.

\subsubsection{Data Preparation}
The main sources include curriculum standards, examination syllabi, training programs, vocational qualification syllabi, and authoritative textbooks. These materials define stable content boundaries, specify stage-level knowledge and ability requirements, and follow relatively stable revision cycles.

We collect 125 official documents across four education stages. For compulsory education, we use the Compulsory Education Curriculum Plan and Curriculum Standards, together with Core Competencies and Values for Chinese Students' Development. For senior high school, we use the General High School Curriculum Standards. For graduate education, we use the national postgraduate entrance examination syllabi. For vocational qualification, we use qualification examination syllabi from fields such as law, medicine, engineering, and information technology. All documents are converted into plain text and processed through format cleaning, section segmentation, and noise removal.

\subsubsection{Concept Extraction and Standardization}

Rule based screening is first applied to extract candidate teaching concepts. Regular expressions are designed according to the hierarchical structure of curriculum standards and examination syllabi. Section hierarchies are then used to establish preliminary parent child relations among candidates.

The candidates are further refined with Qwen-2.5-72B-Instruct, which is tuned for educational instructions. The model removes non-conceptual items such as teaching methods, assessment methods, and organizational forms. It also checks concept granularity by splitting overly broad concepts and merging overly fragmented procedural items. Finally, KnowledgePoint names are normalized with an education terminology dictionary covering all stages. Each KnowledgePoint is assigned a concise definition based on the original standards to improve semantic consistency.

\subsubsection{Relation Construction}

Initial semantic relations are constructed from document structure and terminology resources. Parent child relations are extracted from section hierarchies. Prerequisite relations are identified from temporal and progression expressions in teaching requirements. Lateral relations are extracted from parallel concepts within the same section. Synonym relations are obtained from the standardized terminology dictionary.

\subsection{Knowledge Mapping}

A single extraction method cannot fully balance precision and recall, and LLM-generated knowledge may contain hallucinations. We therefore design a multi-agent mapping framework to improve mapping accuracy and support error correction. The framework consists of five agents: Retrieval Agent, Proposer Agent, Queryer Agent, Verification Agent, and Recovery Agent.

\paragraph{Retrieval Agent.}
The Retrieval Agent retrieves candidate KnowledgePoints for each educational resource segment. It combines exact matching and semantic retrieval. Exact matching uses keywords and the terminology dictionary to identify KnowledgePoints with identical names or highly similar aliases. Semantic retrieval uses BGE-M3 to encode resource segments and KnowledgePoint definitions into the same vector space. The top-3 candidates are retrieved by cosine similarity. The two result sets are deduplicated, reranked, and fused with source weights to form the final candidate set.

\paragraph{Proposer Agent.}
The Proposer Agent generates initial mapping proposals from the candidate set. It considers the resource segment, candidate definitions, subject, education stage, and contextual evidence. Each proposal contains the target KnowledgePoint, relation type, confidence score, and supporting evidence. Evidence may include key terms, definition fragments, curriculum standard entries, or section context.

\paragraph{Queryer Agent.}
The Queryer Agent reviews each proposal from an adversarial perspective. It checks semantic consistency, relation validity, and evidence sufficiency. Semantic consistency examines whether the resource segment falls within the definition of the candidate KnowledgePoint. Relation validity checks whether the proposed relation follows ontology constraints and educational logic. Evidence sufficiency checks whether the mapping is supported by reliable evidence rather than surface lexical similarity. When a problem is found, the Queryer Agent asks the Proposer Agent to revise or justify the proposal.

\paragraph{Verification Agent.}
The Verification Agent coordinates proposal review and makes the final decision. The maximum number of interaction rounds is set to three. In each round, the Verification Agent sends the Queryer Agent's feedback to the Proposer Agent, which may add evidence, adjust candidate ranking, or revise the relation type. If the agents agree on semantic scope, relation type, and evidence, the mapping is accepted and written into the graph. Otherwise, the sample is marked as failed and passed to the Recovery Agent.

\paragraph{Recovery Agent.}
The Recovery Agent repairs failed samples through expanded retrieval, clustering, and structural revision. For failures caused by missing candidates, retrieval is expanded with alias extension, parent concept recall, and cross-source semantic search. For failures caused by vague definitions, the agent checks whether nodes should be merged or definitions refined. For failures caused by unclear relation boundaries, the error is traced to relation recognition, node granularity, or ontology design. For long-tail KnowledgePoints with limited evidence or overly fine granularity, the agent clusters them with adjacent or higher-level nodes and merges them when appropriate to reduce graph fragmentation.

\subsection{Quality Evaluation}

Graph quality is evaluated through automatic screening and expert review. An LLM first checks all nodes, relations, and mappings, and marks samples with low confidence, potential conflicts, or structural anomalies. Experts fully review high-risk samples, while low-risk samples are validated through 5\% random sampling.

\subsection{LLM screening}
GPT-5.4-mini is used for large-scale screening of graph nodes, relations, and mappings. Each sample is scored from 0 to 1 on four dimensions: ontology completeness, definition accuracy, mapping accuracy, and relation consistency. The four dimensions are equally weighted, and their average is used as the final score. Samples below 0.6 are marked for expert review, samples below 0.3 are removed, and samples at or above 0.9 are treated as high-confidence samples.

Ontology completeness checks the coverage of predefined Domain and Subject levels, isolated nodes, unlinked edges, and missing required attributes. Definition accuracy checks whether KnowledgePoint names and definitions follow authoritative sources such as curriculum standards and examination syllabi. Mapping accuracy checks whether each resource segment is correctly linked to the target KnowledgePoint and relation type. Relation consistency checks cyclic prerequisites, contradictory relations, redundant relations, and prerequisite links that conflict with educational cognition.

After screening, removed samples are deleted from the graph, samples marked for review are sent to experts, and high-confidence samples enter sampling validation.

\subsection{Expert review}
We organize a panel of 16 experts, including 10 subject experts, 2 ontology experts, and 4 educational technology experts. Subject experts check factual correctness across basic, higher, and vocational education. Ontology experts review semantic structure and ontology compliance. Educational technology experts examine resource mapping and data consistency. Before formal review, all experts complete guideline training and annotate 100 pilot samples to align their criteria.

The review follows a double-blind independent process. Each flagged sample is reviewed by at least two experts from the same field. Cohen's kappa reaches 0.70 in the pilot annotation, indicating acceptable inter-annotator agreement. We also report the disagreement rate and adjudication rate to describe review stability. A three-member panel, resolve disagreements and unresolved major cases are decided by the lead expert.

Experts review samples from three perspectives. Content accuracy checks names, definitions, semantic relations, and mappings against authoritative documents and educational logic. System completeness checks whether the graph covers required knowledge and key competencies, including prerequisite and interdisciplinary relations. Ontology compliance checks required attributes, relation types, terminology consistency, and logical constraints.

Based on expert feedback, the knowledge graph is revised by correcting, removing, merging, or decomposing nodes and relations. The final statistics of WoW-EKG are shown in Table~\ref{tab:taxonomy_stats}.

\begin{table*}[htbp]
    \centering
    \caption{Educational domain taxonomy and the statistics of \textsc{WoW-EKG}. $|V|$: KnowledgePoints (including one fallback ``Other'' node per subject); $|E|$: edges, counting the subject's \texttt{HAS\_SUBJECT} and \texttt{HAS\_KP} edges and the semantic relations that start from its KnowledgePoints; $|C|$: instruction instances mapped to its KnowledgePoints. ``--'': defined in the taxonomy but not yet instantiated.}
    \label{tab:taxonomy_stats}
    \resizebox{\textwidth}{!}{
    \begin{tabular}{llrrr}
        \toprule
        \multicolumn{1}{c}{\textbf{Domain}} & \multicolumn{1}{c}{\textbf{Subjects}} & \textbf{Nodes ($|V|$)} & \textbf{Edges ($|E|$)} & \textbf{Chunks ($|C|$)} \\
        \midrule

        \rowcolor{gray!15} \multicolumn{5}{c}{\textit{\textbf{Formal Disciplinary Education (Narrow Sense)}}} \\
        \midrule
        \multirow{7}{*}{Computer Science \& Technology} & Theoretical Computer Science & 102 & 534 & 3,166 \\
         & Computer Architecture & 420 & 2,120 & 12,203 \\
         & Computer Software & 1,605 & 7,149 & 66,923 \\
         & Computer Networks & 835 & 3,811 & 12,791 \\
         & Artificial Intelligence & 114 & 539 & 7,683 \\
         & Computer Applications & 865 & 3,664 & 32,373 \\
         & Others & -- & -- & -- \\
        \midrule
        \multirow{11}{*}{Engineering \& Technology} & Mechanical Engineering & 266 & 1,060 & 798 \\
         & Electrical \& Electronics Engineering & 272 & 1,023 & 795 \\
         & Control Science \& Engineering & 45 & 212 & 332 \\
         & Civil \& Architectural Engineering & 3,448 & 13,077 & 2,337 \\
         & Chemical \& Materials Engineering & 62 & 205 & 91 \\
         & Optical Engineering \& Instrumentation & -- & -- & -- \\
         & Energy \& Power Engineering & 19 & 59 & 10 \\
         & Information \& Communication Engineering & 757 & 3,177 & 700 \\
         & Aerospace \& Transportation Engineering & 22 & 75 & 29 \\
         & Environmental Science \& Engineering & 1,012 & 3,655 & 1,572 \\
         & Others & 1,259 & 5,092 & 1,566 \\
        \midrule
        \multirow{6}{*}{Economics \& Management} & Theoretical Economics & 14 & 54 & 17 \\
         & Applied Economics & 1,834 & 6,890 & 1,939 \\
         & Business Administration & 1,482 & 5,782 & 1,209 \\
         & Management Science \& Engineering & 435 & 1,813 & 408 \\
         & Public Administration & 8 & 19 & 9 \\
         & Others & 3 & 7 & 45 \\
        \midrule
        \multirow{6}{*}{Humanities \& Social Sciences} & Linguistics & 3,455 & 14,592 & 799 \\
         & Education & 2,626 & 11,048 & 1,755 \\
         & Literature & 1,656 & 6,484 & 1,054 \\
         & Philosophy \& History & 2,406 & 9,744 & 3,674 \\
         & Sociology \& Political Science & 1,309 & 5,245 & 2,473 \\
         & Others & 71 & 229 & 103 \\
        \midrule
        \multirow{6}{*}{Natural Sciences \& Technology} & Mathematics & 4,203 & 21,500 & 3,724 \\
         & Physics & 1,549 & 7,098 & 2,179 \\
         & Chemistry & 3,627 & 15,709 & 1,441 \\
         & Biology & 3,543 & 14,558 & 2,018 \\
         & Astronomy \& Earth Sciences & 995 & 4,137 & 476 \\
         & Others & 388 & 1,674 & 198 \\
        \midrule
        \multirow{4}{*}{Law \& Institutional Studies} & Legal Theory & 38 & 147 & 45 \\
         & Specific Legal Branches & 2,819 & 11,989 & 5,520 \\
         & Legal Systems \& Practice & 7,280 & 33,539 & 8,523 \\
         & Others & -- & -- & -- \\
        \midrule
        \multirow{5}{*}{Medical \& Health Sciences} & Clinical Medicine & 7,376 & 28,225 & 449 \\
         & Public Health \& Preventive Medicine & 3,535 & 11,669 & 942 \\
         & Pharmacy & 125 & 425 & 10 \\
         & Nursing \& Rehabilitation Medicine & 24 & 91 & 2 \\
         & Others & 8 & 31 & 1 \\
        \midrule

        \rowcolor{gray!15} \multicolumn{5}{c}{\textit{\textbf{General \& Social Education (Broad Sense)}}} \\
        \midrule
        \multirow{6}{*}{Social Edu. \& Literacy} & General \& Comprehensive Literacy Education & 2,998 & 12,732 & 850 \\
         & Social-Emotional \& Ethical Education & 1,402 & 5,424 & 1,170 \\
         & Cognitive Skills \& Learning Literacy & 2,127 & 9,137 & 572 \\
         & Life Practices \& Social Adaptation & 3,882 & 15,785 & 1,179 \\
         & Cultural Aesthetics \& Classical Literacy & 3,994 & 16,943 & 2,525 \\
         & Others & 6 & 19 & 2 \\
        \midrule
        \multirow{4}{*}{Unclassified \& General KB} & Comprehensive Reference Materials & 8 & 19 & 1 \\
         & General Tools \& Methods & 5 & 15 & 0 \\
         & Interdisciplinary \& Unclassified Content & 30 & 103 & 12 \\
         & Others & -- & -- & -- \\
        \midrule
        \multicolumn{2}{l}{\textbf{Total} (9 domains, 57 subjects)} & \textbf{76,364} & \textbf{318,328} & \textbf{188,693} \\
        \bottomrule
    \end{tabular}
    }
\end{table*}

\section{Related Works}
\label{app:related}

Table~\ref{tab:dataset_comparison_main} compares WoW with representative training corpora, instruction datasets, and evaluation benchmarks along three dimensions: how the data is organized (\textit{Fragmentation and Organization}), what supervision it provides (\textit{Supervision and Quality}), and which stage of model development it serves (\textit{Model Utility}). We discuss each group below and then position WoW.

\paragraph{Pre-training corpora.}
Large general-purpose corpora are assembled from mixtures of web and curated sources: ROOTS \citep{laurenccon2022bigscience} spans 1.6\,TB in many languages, Dolma \citep{soldaini2024dolma} provides 3T tokens for open pre-training research, and DataComp-LM \citep{li2024datacomp} turns corpus construction into a controlled benchmark over a 240T-token pool. A second line filters web text for educational value. FineWeb-Edu \citep{penedo2024fineweb} scores English pages with an educational-quality classifier and retains 1.3T tokens; Chinese FineWeb-Edu V2 \citep{yu2025opencsg} and ArabicWeb-Edu \citep{hawasly2025arabicweb} apply the same idea to Chinese and Arabic. A third line builds domain corpora: S2ORC \citep{lo2020s2orc} for academic papers, The Stack \citep{kocetkov2022stack} for source code, Pile of Law \citep{henderson2022pile} and MultiLegalPile \citep{niklaus2024multilegalpile} for legal text, InfiMM-WebMath-40B \citep{han2024infimm} for interleaved mathematical web pages, Biomed-Enriched \citep{touchent2026biomed} and BIOMEDICA \citep{lozano2025biomedica} for biomedical text and figures, and ChemPile \citep{mirza2026chempile} for chemistry. These corpora differ greatly in scope, but they share an organizing principle: data is grouped by \emph{source} (a website, a repository, a jurisdiction, a programming language) or ranked by a \emph{quality score}. Neither tells a practitioner which concepts a corpus covers, how those concepts depend on one another, or which parts of a curriculum are missing. Most of these corpora also provide raw text without explicit supervision, so they serve pre-training but not instruction tuning or evaluation.

\paragraph{Instruction and supervision data.}
Instruction datasets add explicit supervision but are typically organized by task. Super-NaturalInstructions \citep{wang2022super} collects 1,616 NLP tasks with declarative definitions, and MatSci-NLP \citep{song2023matsci} frames materials-science text as seven NLP tasks. In the natural sciences, SMolInstruct \citep{yu2024llasmol} and Mol-Instructions \citep{fang2024mol} provide millions of molecular and biomolecular instructions grouped into task families, and SciInstruct \citep{zhang2024sciinstruct} contributes 254K scientific-reasoning instructions organized by field and source subject. PIXIU \citep{NEURIPS2023_6a386d70} pairs financial instructions with an evaluation suite. Closest to our setting, EduChat \citep{dan2023educhat} builds a Chinese educational assistant from an educational corpus and instructions, and Book2Dial \citep{wang2024book2dial} grounds teacher--student dialogues in textbooks. These resources show that domain- and textbook-grounded supervision is valuable, but each covers a single domain or interaction format, and none links its instances to a shared model of educational knowledge across stages and disciplines.

\paragraph{Evaluation benchmarks.}
Broad-coverage exams measure knowledge across many subjects. MMLU \citep{hendrycks2020measuring} covers 57 subjects in English, and C-Eval \citep{huang2023ceval}, CMMLU \citep{li2023cmmlu}, and M3KE \citep{liu2023m3ke} provide Chinese counterparts organized by subject and education level. EXAMS \citep{hardalov2020exams} and M3Exam \citep{zhang2023m3exam} extend school exams to many languages, RACE \citep{lai2017race} tests reading comprehension on middle- and high-school exams, and E-EVAL \citep{hou2024eeval} targets Chinese K--12 education. Multimodal benchmarks such as MMMU \citep{yue2024mmmu}, CMMMU \citep{zhang2024cmmmu}, EXAMS-V \citep{das2024examsv}, ScienceQA \citep{lu2022scienceqa}, and ChartQA \citep{masry2022chartqa} add images, diagrams, and charts to disciplinary questions. A large group of benchmarks probes individual disciplines in depth. In the natural sciences, these include SciBench \citep{wang2023scibench} and SciEval \citep{sun2024scieval} for college-level and research-oriented science; ChemBench \citep{mirza2024large}, ChemVLM/MMChemBench \citep{li2025chemvlm}, and MoleculeQA \citep{lu2024moleculeqa} for chemistry; OlympiadBench \citep{he2024olympiadbench}, PhysUniBench \citep{wang2025physunibench}, UGPhysics \citep{xu2025ugphysics}, SeePhys \citep{xiang2026seephys}, and PhysReason \citep{zhang2025physreason} for physics; OlympicArena \citep{huang2024olympicarena} for olympiad problems across fields; and LAB-Bench \citep{laurent2024lab} for biology research tasks. Other domains are covered by MedMCQA \citep{pal2022medmcqa} for medicine, LegalBench \citep{guha2023legalbench} for law, TAT-QA \citep{zhu2021tat} and FinQA \citep{chen2021finqa} for financial reasoning over tables and text, QASPER \citep{dasigi2021dataset} for questions over scientific papers, and SWE-bench \citep{jimenez2024swe} for software engineering. Benchmarks on values and argumentation, such as ETHICS \citep{hendrycks2020aligning}, Moral Stories \citep{emelin2021moral}, Social-Chem-101 \citep{forbes2020social}, and CEAMC \citep{ren2024ceamc}, assess literacy-oriented competence beyond disciplinary knowledge. These benchmarks are indispensable for measuring capabilities, but most provide no training split, and their taxonomies (subject labels, exam levels, task types) are defined independently for each benchmark, so results cannot be traced back to the training data that a model has seen.

\paragraph{Knowledge-structured educational resources.}
A few resources organize educational content around explicit knowledge points. CK12 \citep{you2024ck12} builds a Chinese K--12 benchmark on a knowledge graph of 2,573 knowledge points; MDK12-Bench \citep{zhou2026mdk12} annotates 140K multimodal K--12 exam instances with 6,827 knowledge points; and K12-KGraph \citep{liang2026k12} constructs a curriculum-aligned graph of 10,685 nodes and 23,278 edges together with a science benchmark and a small training set. These works demonstrate that a knowledge structure makes coverage measurable and evaluation interpretable. However, they are confined to K--12 education and mostly to the natural sciences, and the graph is used to index evaluation items rather than to drive the collection and construction of training data at scale.

\paragraph{Positioning of WoW.}
WoW differs from prior resources along all three dimensions of Table~\ref{tab:dataset_comparison_main}. In organization, an Educational Knowledge Graph with 76K KnowledgePoints and 318K typed edges across nine domains and four education stages serves as the taxonomic backbone (Appendix~\ref{app:kg}); it is about seven times larger in nodes and fourteen times larger in edges than K12-KGraph, the largest curriculum graph in the table. Every instance is mapped to a KnowledgePoint, so coverage can be measured and gaps can be filled systematically. In supervision, WoW provides problems with reference answers that are grounded in evidence retrieved from their source documents and filtered by explicit quality criteria (\S\ref{sec:method}), whereas most corpora in the table offer either raw text or task-specific labels. In utility, WoW supports training as well as evaluation, with document-level provenance linking the two, so that a model's behavior can be traced back to the knowledge it was trained on. Across all rows of the table, WoW is the only resource that combines a knowledge-graph backbone, multi-disciplinary coverage beyond K--12, and support for both training and evaluation.

\begin{table*}[p]
\centering
\caption{Comparison of representative training-oriented corpora and evaluation benchmarks.}
\label{tab:dataset_comparison_main}

\vspace{-1mm}
\begingroup
\fontsize{5.35pt}{6.55pt}\selectfont
\setlength{\tabcolsep}{1.10pt}
\renewcommand{\arraystretch}{1.22}

\begin{threeparttable}
\begin{adjustbox}{max width=\textwidth,center}
\begin{tabular}{@{}l l c l l c c c c c@{}}
\toprule
& \multicolumn{5}{c}{\textbf{Fragmentation and Organization}}
& \multicolumn{2}{c}{\textbf{Supervision and Quality}}
& \multicolumn{2}{c}{\textbf{Model Utility}} \\
\cmidrule(lr){2-6}
\cmidrule(lr){7-8}
\cmidrule(lr){9-10}
\textbf{Dataset}
& \textbf{Scale}
& \textbf{Domain}
& \textbf{Scope / Form}
& \textbf{Taxonomy Backbone}
& \textbf{Lang.}
& \textbf{MM}
& \textbf{Sup.}
& \textbf{Train}
& \textbf{Eval.} \\
\midrule

\multicolumn{10}{@{}l}{\textbf{\textit{Training Corpora \& Instruction Data}}} \\
\midrule

S2ORC~\cite{lo2020s2orc}
& 81.1M papers; 8.1M full texts
& MultiDisc
& Academic papers
& Paper metadata
& EN
& \no
& \no
& \yes
& \no \\

The Stack~\cite{kocetkov2022stack}
& 3.1TB code
& CS
& Source code
& Language
& Code
& \no
& \no
& \yes
& \no \\

Pile of Law~\cite{henderson2022pile}
& 256GB
& LawInst
& Legal text
& Legal source
& EN
& \no
& \no
& \yes
& \no \\

Super-NaturalInstructions~\cite{wang2022super}
& 1,616 tasks
& CS
& Instruction tasks
& Task definition
& EN
& \no
& \yes
& \yes
& \yes \\

ROOTS~\cite{laurenccon2022bigscience}
& 1.6TB corpus
& MultiDisc
& Web / text
& Source mix
& Multi
& \no
& \no
& \yes
& \no \\

MatSci-NLP~\cite{song2023matsci}
& 33K abstracts; 7 NLP tasks
& EngTech
& Materials science text
& NLP task type
& EN
& \no
& \yes
& \yes
& \no \\

EduChat~\cite{dan2023educhat}
& Edu corpus; inst.
& PanEdu
& Edu chatbot
& Prompt / instruction
& ZH
& \no
& \yes
& \yes
& \yes \\

MultiLegalPile~\cite{niklaus2024multilegalpile}
& 689GB; 24 lang.; 17 juris.
& LawInst
& Legal corpus
& Jurisdiction / source
& Multi
& \no
& \no
& \yes
& \no \\

SWE-bench~\cite{jimenez2024swe}
& 2,294 issue-pull pairs
& EngTech
& Software engineering
& GitHub repo / issue
& EN
& \no
& \yes
& \no
& \yes \\

Dolma~\cite{soldaini2024dolma}
& 3T tok.
& MultiDisc
& General pretrain
& Source mix
& EN
& \no
& \no
& \yes
& \no \\

FineWeb-Edu~\cite{penedo2024fineweb}
& 1.3T tok.
& PanEdu
& Edu web
& Quality score
& EN
& \no
& \no
& \yes
& \no \\

DataComp-LM~\cite{li2024datacomp}
& 240T tok. pool; 2.6T baseline
& MultiDisc
& Web pretrain
& Data recipe
& EN
& \no
& \no
& \yes
& \yes \\

InfiMM-WebMath-40B~\cite{han2024infimm}
& 24M pages; 40B tok.; 85M imgs.
& NatTech
& Interleaved web
& Page-level interleave
& EN/ZH
& \yes
& \no
& \yes
& \no \\

Book2Dial~\cite{wang2024book2dial}
& Textbook-grounded dial.
& PanEdu
& Textbook dialogue
& Textbook grounding
& EN
& \no
& \yes
& \yes
& \yes \\

Chinese FineWeb-Edu V2~\cite{yu2025opencsg}
& 188M texts; 420B tok.
& PanEdu
& Chinese edu web
& Quality score
& ZH
& \no
& \no
& \yes
& \no \\

ArabicWeb-Edu~\cite{hawasly2025arabicweb}
& 1M labeled docs; 800B tok. exp.
& PanEdu
& Arabic edu web
& Quality class
& AR
& \no
& \no
& \yes
& \no \\

Biomed-Enriched~\cite{touchent2026biomed}
& 400K labeled paras.; PMC-OA scale
& MedHealth
& Biomedical text
& Quality / domain label
& EN
& \no
& \yes
& \yes
& \no \\

BIOMEDICA~\cite{lozano2025biomedica}
& 6M+ articles; 24M image-text pairs
& MedHealth
& Biomedical image-text
& Article / figure
& EN
& \yes
& \yes
& \yes
& \no \\

ChemPile~\cite{mirza2026chempile}
& 255GB; 76.7B tok.; 260M docs.
& NatTech
& Chemistry foundation corpus
& 7 content families
& EN
& \yes
& \yes
& \yes
& \yes \\

SMolInstruct~\cite{yu2024llasmol}
& 3.3M pairs; 1.6M molecules; 14 tasks
& NatTech
& Molecular instructions
& 4 families / 14 tasks
& EN
& \no
& \yes
& \yes
& \yes \\

Mol-Instructions~\cite{fang2024mol}
& 2,043,587 inst.; 17 tasks
& NatTech
& Biomolecular instructions
& 3 components / 17 tasks
& EN
& \no
& \yes
& \yes
& \yes \\

SciInstruct~\cite{zhang2024sciinstruct}
& 254,051 inst.; 123,869 phys./chem.
& NatTech
& Scientific reasoning inst.
& Field / source subject
& Multi
& \no
& \yes
& \yes
& \no \\

\midrule
\multicolumn{10}{@{}l}{\textbf{\textit{Evaluation Benchmarks}}} \\
\midrule

RACE~\cite{lai2017race}
& 28K passages; 100K Qs
& PanEdu
& Middle/high exams
& Passage / question
& EN
& \no
& \yes
& \yes
& \yes \\

EXAMS~\cite{hardalov2020exams}
& 24K+ Qs; 16 lang.; 24 subj.
& PanEdu
& High-school exams
& Exam / subject
& Multi
& \no
& \yes
& \no
& \yes \\

Social-Chem-101~\cite{forbes2020social}
& 292K rules; 4.5M ann.
& PanEdu
& Social norms
& Norm dimension
& EN
& \no
& \yes
& \yes
& \yes \\

MMLU~\cite{hendrycks2020measuring}
& 15.9K Qs; 57 subj.
& MultiDisc
& Multi-task exam
& Subject label
& EN
& \no
& \yes
& \no
& \yes \\

ETHICS~\cite{hendrycks2020aligning}
& 130K+ examples
& PanEdu
& Ethical reasoning
& Value category
& EN
& \no
& \yes
& \yes
& \yes \\

Moral Stories~\cite{emelin2021moral}
& Structured moral narratives
& PanEdu
& Moral reasoning
& Narrative schema
& EN
& \no
& \yes
& \yes
& \yes \\

QASPER~\cite{dasigi2021dataset}
& 5,049 Qs; 1,585 papers
& CS
& Scientific papers
& Paper / evidence
& EN
& \no
& \yes
& \yes
& \yes \\

TAT-QA~\cite{zhu2021tat}
& Financial reports QA
& EconMgmt
& Table-text QA
& Table / text
& EN
& \no
& \yes
& \yes
& \yes \\

FinQA~\cite{chen2021finqa}
& Financial QA; gold programs
& EconMgmt
& Numerical QA
& Program / report
& EN
& \no
& \yes
& \yes
& \yes \\

ScienceQA~\cite{lu2022scienceqa}
& 21,208 QA
& NatTech
& K--12 science QA
& QA / rationale
& EN
& \yes
& \yes
& \yes
& \yes \\

ChartQA~\cite{masry2022chartqa}
& 9.6K human Qs; 23.1K gen. Qs
& PanEdu
& Chart QA
& Chart / table
& EN
& \yes
& \yes
& \yes
& \yes \\

MedMCQA~\cite{pal2022medmcqa}
& 194K+ MCQs; 21 subj.
& MedHealth
& Medical exams
& Medical subject
& EN
& \no
& \yes
& \yes
& \yes \\

PIXIU~\cite{NEURIPS2023_6a386d70}
& 136K inst.; 4 eval tasks
& EconMgmt
& Finance inst. \& eval
& Financial NLP task
& EN
& \no
& \yes
& \yes
& \yes \\

C-Eval~\cite{huang2023ceval}
& 13,948 MCQs; 52 subj.
& MultiDisc
& Chinese exams
& Subject / level
& ZH
& \no
& \yes
& \no
& \yes \\

CMMLU~\cite{li2023cmmlu}
& 11,528 Qs; 67 subj.
& MultiDisc
& Chinese knowledge
& Subject label
& ZH
& \no
& \yes
& \no
& \yes \\

M3KE~\cite{liu2023m3ke}
& 20,477 Qs; 71 tasks
& PanEdu
& Multi-level exams
& Level / task
& ZH
& \no
& \yes
& \no
& \yes \\

M3Exam~\cite{zhang2023m3exam}
& 12,317 Qs; 9 lang.; 3 levels
& PanEdu
& Multilingual exams
& Exam / level
& Multi
& \yes
& \yes
& \no
& \yes \\

LegalBench~\cite{guha2023legalbench}
& 162 tasks
& LawInst
& Legal reasoning
& Legal task
& EN
& \no
& \yes
& \no
& \yes \\

E-EVAL~\cite{hou2024eeval}
& 4,351 MCQs
& PanEdu
& Chinese K--12
& Stage / subject
& ZH
& \no
& \yes
& \no
& \yes \\

CK12~\cite{you2024ck12}
& 39,452 Qs; 2,573 KPs
& PanEdu
& Chinese K--12
& K--12 KG
& ZH
& \no
& \yes
& \no
& \yes \\

MMMU~\cite{yue2024mmmu}
& 11.5K Qs; 30 subj.
& MultiDisc
& College multi-disc.
& Discipline / subject
& EN
& \yes
& \yes
& \no
& \yes \\

CMMMU~\cite{zhang2024cmmmu}
& 12K QA; 30 subj.
& MultiDisc
& Chinese college
& Discipline / subject
& ZH
& \yes
& \yes
& \no
& \yes \\

EXAMS-V~\cite{das2024examsv}
& 20,932 MM MCQs; 11 lang.
& MultiDisc
& Multimodal exams
& Discipline
& Multi
& \yes
& \yes
& \no
& \yes \\

CEAMC~\cite{ren2024ceamc}
& Essay corpus
& HumSoc
& Chinese essay
& Argument component
& ZH
& \no
& \yes
& \yes
& \yes \\

ChemBench~\cite{mirza2024large}
& $>$2,700 QA pairs
& NatTech
& Chemistry knowledge / reasoning
& Topic / skill / difficulty
& EN
& \no
& \yes
& \no
& \yes \\

ChemVLM/MMChemBench~\cite{li2025chemvlm}
& 200K train QA; 3K test
& NatTech
& Multimodal chemistry QA
& Chemical task type
& EN/ZH
& \yes
& \yes
& \yes
& \yes \\

MoleculeQA~\cite{lu2024moleculeqa}
& 61,574 QA; 23K molecules
& NatTech
& Molecular 4-way QA
& 4 aspects
& EN
& \no
& \yes
& \yes
& \yes \\

OlympiadBench (Physics)~\cite{he2024olympiadbench}
& 2,334 physics Qs; 1,958 imgs.
& NatTech
& Olympiad physics
& Topic / problem type
& EN/ZH
& \yes
& \yes
& \no
& \yes \\

PhysUniBench~\cite{wang2025physunibench}
& 3,304 Qs; 8 subdisciplines
& NatTech
& Undergraduate physics QA
& Subdisciplines / difficulty
& EN/ZH
& \yes
& \yes
& \no
& \yes \\

UGPhysics~\cite{xu2025ugphysics}
& 5,520 problems; 13 subjects
& NatTech
& Undergraduate physics QA
& 3 domains / 13 subj. / 59 topics
& EN/ZH
& \no
& \yes
& \no
& \yes \\

SeePhys~\cite{xiang2026seephys}
& 2,000 Qs
& NatTech
& Visual physics reasoning
& 7 domains / 21 diagram categories
& EN/ZH
& \yes
& \yes
& \no
& \yes \\

PhysReason~\cite{zhang2025physreason}
& 1,200 problems
& NatTech
& Multimodal physics reasoning
& Knowledge / reasoning / difficulty
& ZH/EN/HI/RU
& \yes
& \yes
& \no
& \yes \\

MDK12-Bench~\cite{zhou2026mdk12}
& 140K instances; 6,827 knowledge-point annotations
& NatTech
& Multimodal K--12 exams
& 6 disciplines
& EN
& \yes
& \yes
& \no
& \yes \\

OlympicArena~\cite{huang2024olympicarena}
& 11,163 problems
& NatTech
& Multi-discipline olympiad
& 7 fields / 62 competitions
& EN/ZH
& \yes
& \yes
& \no
& \yes \\

LAB-Bench~\cite{laurent2024lab}
& 2,457 Qs; 531 MM
& NatTech
& Biology research tasks
& 8 task categories
& EN
& \yes
& \yes
& \no
& \yes \\

K12-KGraph/K12-Bench/K12-Train~\cite{liang2026k12}
& 10,685 nodes; 23,278 edges; 23,640 Qs; 2,267 inst.
& NatTech
& Curriculum KG / science QA
& 7-node / 9-edge curriculum KG
& ZH
& \no
& \yes
& \yes
& \yes \\

SciBench~\cite{wang2023scibench}
& 972 problems
& NatTech
& College science problems
& 3 fields / 10 textbooks
& EN
& \yes
& \yes
& \no
& \yes \\

SciEval~\cite{sun2024scieval}
& 15,901 Qs
& NatTech
& Scientific research ability
&  4 levels / 3 fields
& EN
& \no
& \yes
& \no
& \yes \\

\midrule
\rowcolor{gray!15}
\textbf{WoW (Ours)}
& \textbf{188K inst.; 146M tok.; 76K KPs; 318K edges}
& \textbf{Comprehensive}
& \textbf{Full-cycle edu data}
& \textbf{EKG backbone}
& \textbf{ZH/EN}
& \textbf{\yes}
& \textbf{\yes}
& \textbf{\yes}
& \textbf{\yes} \\

\bottomrule
\end{tabular}
\end{adjustbox}

\end{threeparttable}
\endgroup

\par\vspace{0.5mm}
\noindent
\begin{minipage}{\textwidth}
\footnotesize
\raggedright
\textbf{Note:}
CS = Computer Science \& Technology;
EngTech = Engineering \& Technology;
EconMgmt = Economics \& Management;
HumSoc = Humanities \& Social Sciences;
NatTech = Natural Sciences \& Technology;
LawInst = Law \& Institutional Studies;
MedHealth = Medical \& Health Sciences;
PanEdu = Social Edu. \& Literacy;
MultiDisc = multi-disciplinary knowledge.
\textit{Scope / Form} describes the data form or usage scenario,
while \textit{Taxonomy Backbone} describes the organizing principle.
MM = multimodal data;
Sup. = explicit supervision signals such as answers, rationales,
explanations, programs, dialogues, argument labels, quality labels,
or expert annotations.
Train = support for pre-training, fine-tuning, instruction tuning,
or supervised training splits;
Eval. = benchmark-style evaluation support.
KPs = knowledge points;
tok. = tokens;
Qs = questions;
ann. = annotations;
paras. = paragraphs;
abs. = abstracts.
\yes\ denotes explicit support and \no\ denotes no explicit support.
\par
\end{minipage}

\par\vspace{-1mm}
\end{table*}

\end{document}